\documentclass[12pt]{article}
\usepackage{amsmath}
\usepackage{graphicx}
\usepackage{natbib}
\usepackage{url} 
\usepackage{amssymb}
\usepackage{amsthm}
\usepackage{caption}
\usepackage{setspace}

\usepackage{booktabs}
\usepackage{bbm}

\usepackage{hyperref}
\hypersetup{colorlinks,citecolor=blue,urlcolor=blue,linkcolor=blue}

\newcommand{\e}{\epsilon}
\newcommand{\E}{\mathbb{E}}
\newcommand{\R}{\mathbb{R}}
\newcommand{\one}{\mathbbm{1}}
\newcommand{\prob}{\mathbb{P}}
\DeclareMathOperator{\Var}{Var}

\DeclareMathOperator{\argmin}{argmin}
\newcommand{\p}[1]{\left(#1\right)}
\newcommand{\Norm}[1]{\left\lVert#1\right\rVert}

\newtheorem{theorem}{Theorem}
\newtheorem{assumption}{Assumption}

\newcommand\indep{\protect\mathpalette{\protect\independenT}{\perp}}
\def\independenT#1#2{\mathrel{\rlap{$#1#2$}\mkern2mu{#1#2}}}

\begin{document}

\def\spacingset#1{\renewcommand{\baselinestretch}%
{#1}\small\normalsize} \spacingset{1}


\title{Local Linear Forests}
\author{Rina Friedberg\thanks{
     R.F. was supported by the DoD, Air Force Office of Scientific Research, National Defense Science and Engineering Graduate (NDSEG) Fellowship, 32 CFR 168a. The authors also gratefully acknowledge support by the Sloan Foundation, ONR grant N00014-17-1-2131, and NSF grant DMS-1916163. S. W. was supported by a Facebook Faculty Award. The authors would like to thank Guido Imbens, Art Owen, Evan Rosenman, and Steve Yadlowsky for useful comments and discussion. R.F. is currently at LinkedIn, and this paper was included as part of her PhD dissertation at Stanford's Statistics Department.}\\ \texttt{rfriedberg@linkedin.com} \and Julie Tibshirani \\ \texttt{julietibs@gmail.com} \and Susan Athey \\ \texttt{athey@stanford.edu} \and  Stefan Wager \\ \texttt{swager@stanford.edu}}
\maketitle
  \maketitle

\bigskip
\begin{abstract}
\onehalfspacing
Random forests are a powerful method for non-parametric regression, but are limited in their ability to fit smooth signals. Taking the perspective of random forests as an adaptive kernel method, we pair the forest kernel with a local linear regression adjustment to better capture smoothness. The resulting procedure, \textit{local linear forests}, enables us to improve on asymptotic rates of convergence for random forests with smooth signals, and provides substantial gains in accuracy on both real and simulated data. We prove a central limit theorem valid under regularity conditions on the forest and smoothness constraints, and propose a computationally efficient construction for confidence intervals. Moving to a causal inference application, we discuss the merits of local regression adjustments for heterogeneous treatment effect estimation, and give an example on a dataset exploring the effect word choice has on attitudes to the social safety net. Last, we include simulation results on real and generated data. A software implementation is available in the R package grf.
\end{abstract}

\noindent%
\vfill

\newpage
\spacingset{1.5} 

\section{Introduction}\label{sec-introduction}
Random forests \citep*{breiman2001} are a popular method for non-parametric regression that have proven effective across many application areas \citep{cutler2007random,diaz2006gene,svetnik2003random}.
A major weakness of random forests, however, is their inability to exploit smoothness in the regression surface they are estimating. As an example, consider the following setup:
We simulate $X_1, \dots, X_n$ independently
from the uniform distribution on $[0,1]^{20}$, with responses  
\begin{equation}\label{boundary-eq}
y_i = \log\left(1+\exp(6 X_{i1})\right) + \e, ~~ \e \sim \mathcal{N}(0, \, 20),
\end{equation}
and our goal is to estimate $\mu({x_0}) = \mathbb{E}[Y \,|\, X = {x_0}]$.
The left panel of Figure \ref{figure-confidence} shows a set of predictions on this data from a random forest.
The forest is unable to exploit strong local trends and, as a result, fits the target function using
qualitatively the wrong shape: The prediction surface resembles a step function as opposed to a smooth curve.

\begin{figure}[!t]
\begin{tabular}{cc}
\includegraphics[width=0.45\textwidth]{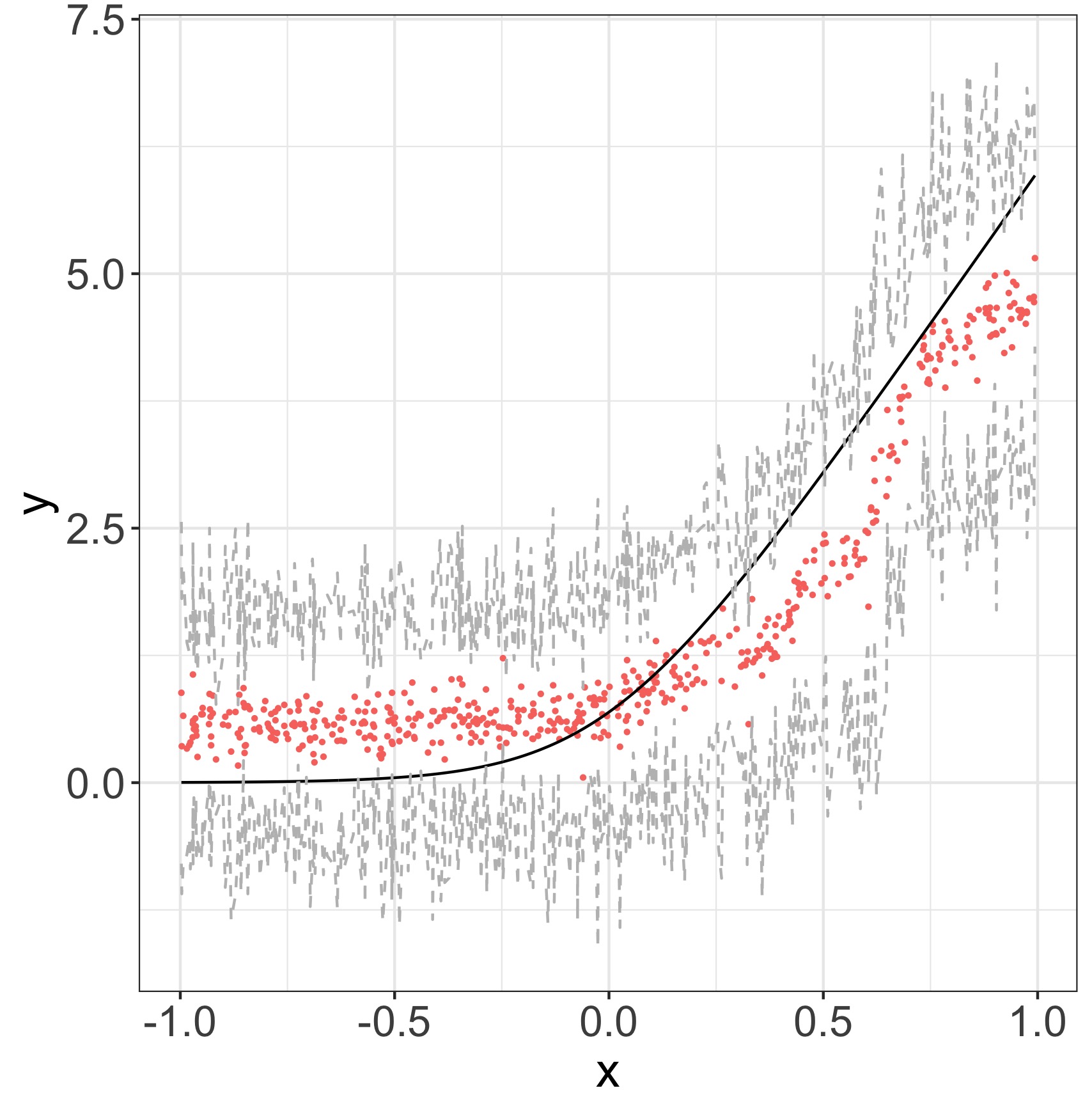} & 
\includegraphics[width=0.45\textwidth]{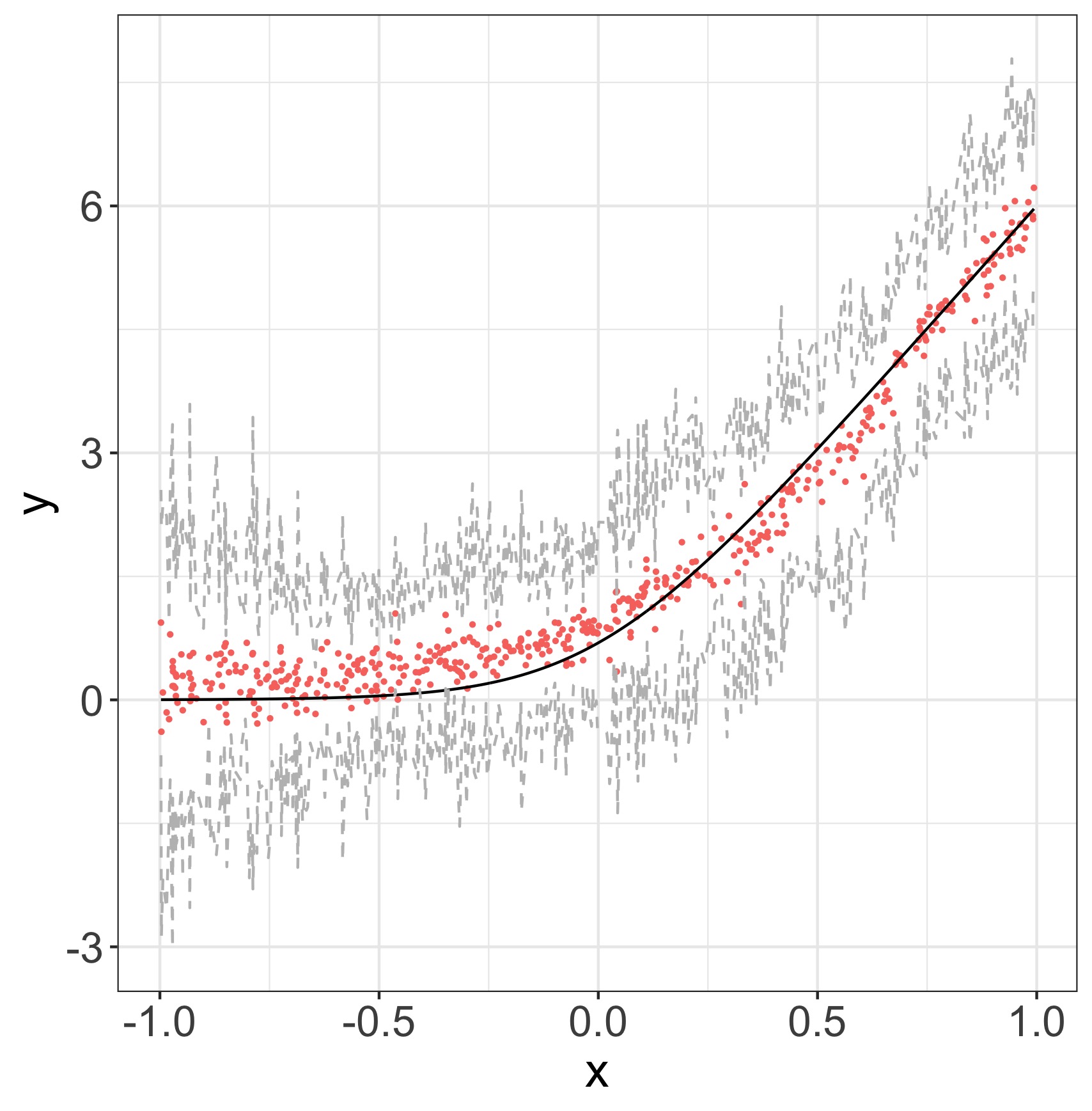} \\
Random forest & Local linear forest \\
\end{tabular}
\caption{
Example 95\% confidence intervals from generalized random forests (left) and local linear forests (right) on out of bag predictions from equation \ref{boundary-eq}.  Training data were simulated from equation (\ref{boundary-eq}), with $n=500$ training points, dimension $d=20$ and errors $\e \sim N(0,20)$. Forests were trained using the  \texttt{R} package \texttt{grf} \citep{package-grf} and tuned via cross-validation.  True signal is shown as a smooth curve, with dots corresponding to forest predictions, and upper and lower bounds of pointwise confidence intervals connected in the dashed lines.}
\end{figure}\label{figure-confidence}

In order to address this weakness, we take the perspective of random forests as an adaptive kernel method. This interpretation follows work by \citet*{athey2016}, \citet*{hothorn}, and \citet*{meinshausen2006quantile}, and complements the traditional view of forests as an ensemble method (i.e., an average of predictions made by individual trees). Specifically, random forest predictions can be written as
\begin{equation}
\label{eq:rf-kernel}
\hat{\mu}_{\text{rf}}({x_0}) = \sum_{i=1}^n \alpha_i({x_0}) \, Y_i, 
\end{equation}
where the weights $\alpha_i({x_0})$, which are defined in the upcoming display \eqref{eq-weights}, encode the weight given by the forest to the
$i$-th training example when predicting at ${x_0}$.
Now, as is well-known in the literature on non-parametric regression,
if we want to fit smooth signals without some form of neighborhood averaging
(e.g., kernel regression, $k$-NN, or matching for causal inference), it is helpful to use
a local regression adjustment to correct for potential misalignment between a test point and its neighborhood
\citep*{abadie2011bias,cleveland1988locally,GVK19282144X,heckman1998matching,
Load:1999,10.2307/3532868,Stone1977,tibshirani1987local}.  These types of adjustments are particularly important near boundaries,
where neighborhoods are asymmetric by necessity. With many covariates, these adjustments are also important away
from boundaries given that  local neighborhoods are often unbalanced due to sampling variation.

The goal of this paper is to improve the accuracy of forests on smooth signals using regression adjustments, potentially in many dimensions.  By using the local regression adjustment, it is possible to adjust for asymmetries and imbalances in the set of nearby points used for prediction, ensuring that the weighted average of the feature vector of neighboring points is approximately equal to the target feature vector, and that predictions are centered.  
The improvement to forests from the regression adjustment is most likely to be large in cases where some features have strong effects with moderate curvature, so that regression adjustments are both effective and important.  

In their simplest form, local linear forests take the forest weights $\alpha_i({x_0})$, and use them for
local regression:
\begin{equation}\label{loss}
\begin{pmatrix} \hat{\mu}({x_0}) \\ \hat{\theta}({x_0}) \end{pmatrix} = \argmin_{\mu,\theta} \left\{\sum_{i=1}^n \alpha_i({x_0}) (Y_i - \mu({x_0}) - (X_i - {x_0})\theta({x_0}) )^2 + \lambda ||\theta({x_0})||_2^2\right\}.
\end{equation}
Here \smash{$\hat{\mu}({x_0})$} estimates the conditional mean function $\mu({x_0})$,
and \smash{$\theta({x_0})$} corrects for the local trend in $X_i- x$.
The ridge penalty $\lambda ||\theta({x_0})||_2^2$ prevents overfitting to the local trend, and plays a key role
both in simulation experiments and asymptotic convergence results.
Then, as discussed in Section \ref{sub-sec-splits}, we can improve the performance of local linear forests by modifying the
tree-splitting procedure used to get the weights $\alpha_i({x_0})$, and making it account for the fact that
we will use local regression to estimate $\mu({x_0})$.
As a first encouraging result, in the motivating example from Figure \ref{figure-confidence},
local linear forests have improved upon the fit of standard forests.

These improvements extend to many other types of forests, such as quantile regression forests \citep*{meinshausen2006quantile} or, more broadly, generalized random forests \citep*{athey2016}. An extension of primary interest is to causal forests as proposed by \citet*{athey2016}, which we discuss in Section \ref{sec-causal} in detail; other cases are analogous.

Our main formal result is a Central Limit Theorem for the predictions $\hat{\mu}({x_0})$ from a local linear forest at a given test point $x$, specifying the asymptotic convergence rate and its dependence on subsampling and smoothness of $\mu({x_0})$. This allows us to build pointwise Gaussian confidence intervals, giving practitioners applicable uncertainty quantification. Observe that in Figure \ref{figure-confidence}, the bias of regression forest predictions affects not only the accuracy prediction curve but also the coverage corresponding confidence intervals, which are not centered on the true function. Local linear forests, in addressing this issue, improve over regression forests in both predictive performance and confidence interval coverage. Strikingly, our local linear forest confidence intervals simultaneously achieve better coverage and are shorter than those built using regression forests.

A simple form of \eqref{loss}, without regularization or modified tree-splitting procedures,
was also considered in a recent paper by \citet*{bloniarz2016supervised}. However, they only report modest performance improvements over basic regression
forests; for example, on the ``Friedman function'' they report roughly a 5\% reduction in mean-squared error.
In contrast, we find fairly large, systematic improvements from local linear forests; see, e.g., Figure
\ref{friedman-sims} for corresponding results on the same Friedman function. It thus appears that our 
algorithmic modifications via regularization and optimized splitting play a qualitatively important
role in getting local linear forests to work well. These empirical findings are also mirrored in our
theory. For example, in order to prove rates of convergence for local linear forests that can exploit
smoothness of $\mu(\cdot)$ and improve over corresponding rates available for regression forests,
we need an appropriate amount of regularization in \eqref{loss}. 

Finally, one can also motivate local linear forests from the starting point of local linear regression.
Despite working well in low dimensions, classical approaches to local linear regression are not applicable to even moderately high-dimensional problems. (This is a well-known problem. The popular core \texttt{R} function \texttt{loess} \citep*{r-core}
allows only 1-4 predictors, while \texttt{locfit} \citep*{locfit} crashes on the simulation from \eqref{boundary-eq} with $d\ge 7$.) 
In contrast, random forests are adept at fitting high-dimensional signals, both in terms of their stability and computational efficiency. 
From this perspective, random forests can be seen as an effective way of producing weights to use in local linear regression. 
In other words, local linear forests aim to combine the adaptivity of random forests 
and the ability of local linear regression to capture smoothness.

An implementation of local linear forests, compliant with the assumptions detailed in Section \ref{sec-theory},
is available in the \texttt{R} package \texttt{grf} \citep{package-grf,r-core}.

\subsection{Empirical Example: Wage Regressions}

To illustrate the promise of local linear forests, we consider the problem of predicting the logarithm of wages as a function of
covariates including years of education, age, race, and gender; this function plays an important role in the study of labor
markets \citep*{RePEc:nbr:nberwo:9732}. This problem has a mix of continuous and
categorical variables, to which tree-based methods are well suited. 
However, wages tend to have a fairly strong and smooth
association with age and education, and we might expect local regression adjustments to help with this.  In addition, the covariate 
space is large relative to the size of publicly available administrative data, and there are moderate to strong correlations among many
of the covariates, making it challenging to obtain accurate predictions in some regions of the covariate space. 

We consider data from the current population survey (CPS), available from the Minnesota Population Center \citep{cps}. These data describe the wages of 114,291 individuals in 2018 (excluding records that do not contain wage data).  To evaluate how model performance varies with sample size, we divide the data into a large test set, which is used to 
evaluate accuracy overall and in specific regions of the covariate space, and training sets of varying sizes.

We compare local linear forests with ordinary least squares, lasso with interaction terms, gradient boosting, Bayesian additive regression trees, and  random forest. For the lasso, random forests, local linear forests and boosting, we chose tuning parameters via cross-validation;
in particular, for local linear forests, we tuned on leaf size and $\lambda$. Moreover, for our method, we did use a local linear correction
for all variables; rather, we only used non-zero $\theta$-coefficient in \eqref{loss} for continuous predictors that had non-zero coefficients
in a pilot lasso regression \citep{Tibshirani94regressionshrinkage} (in general, we have found screening of variables used for a local linear
correction in \eqref{loss} to benefit both the accuracy and computational performance of our approach).
Table \ref{wages-table} compares predictive performance across several methods, showing that local linear forests can provide
a predictive benefit over competing methods.  

\begin{table}[t]
\begin{center}
\footnotesize
\begin{tabular}{@{}lllllllll@{}}
\toprule \toprule
 & $n_{\text{train}}$ && OLS & Lasso & XGB & BART & RF & LLF \\
\cmidrule{2-9} 
& 2,000 && 4.26 (1.43) & 4.34 (0.13) & 1.18 (0.06) & 1.43 (0.07) & 1.19 (0.07) & 1.10 (0.06) \\
& 5,000 && 4.21 (0.12) & 4.15 (0.13) & 1.17 (0.07) & 1.32 (0.08) & 1.15 (0.07) & 1.03 (0.07) \\
& 10,000 && 4.23 (0.12) & 3.98 (0.11) & 1.01 (0.05) & 1.17 (0.07) & 1.04 (0.07) & 0.95 (0.06) \\
& 50,000 && 4.24 (0.13) & 3.98 (0.12) & 0.91 (0.05) & 1.05 (0.07) & 0.98 (0.06) & 0.92 (0.06) \\
\midrule \midrule
 & Avg. $n_{\text{test}}$ && OLS & Lasso & XGB & BART & RF & LLF \\
\cmidrule{2-9}
Extreme ages & 4051 && 1.92 (0.10) & 1.74 (0.09) & 0.48 (0.03) & 0.52 (0.03) & 0.46 (0.04) & 0.44 (0.03) \\
Less sampled races & 3547 && 3.90 (0.15) & 3.71 (0.14) & 1.02 (0.05) & 1.14 (0.07) & 1.01 (0.08) & 0.95 (0.07) \\
Family size $\ge 6$ & 894 && 4.55 (0.39) & 4.32 (0.37) & 0.85 (0.11) & 1.13 (0.14) & 1.06 (0.11) & 0.96 (0.11)  \\ 
\bottomrule \bottomrule 
\end{tabular}
\end{center}
\caption{Mean squared error for predictions of log wages in CPS data, evaluated on a test set with 40,000 observations, for ordinary least squares (OLS), lasso with interaction terms (lasso), gradient boosting (XGB), Bayesian Additive Regression Trees (BART),  random forests (RF), and local linear forests (LLF). We did a train/test split and then performed 100 replications on each method. The standard deviation across replications of the mean square error estimates is shown in parentheses. The top half of the table shows errors on training sets of size $n_{\text{train}}$. The lower panel shows mean square error on sparse regions of the covariate space. We fix a training dataset of size 20,000. For 100 repetitions, we draw a test set of size 40,000 and report errors on the subset of the test set corresponding to the desired condition (e.g. extreme ages). To give a relative sense of the sparsity of these regions, we report the average number of individuals in the test set.}\label{wages-table}
\end{table}

\begin{figure}[!t]
\centering
\begin{tabular}{cc}
\includegraphics[width=0.4\textwidth]{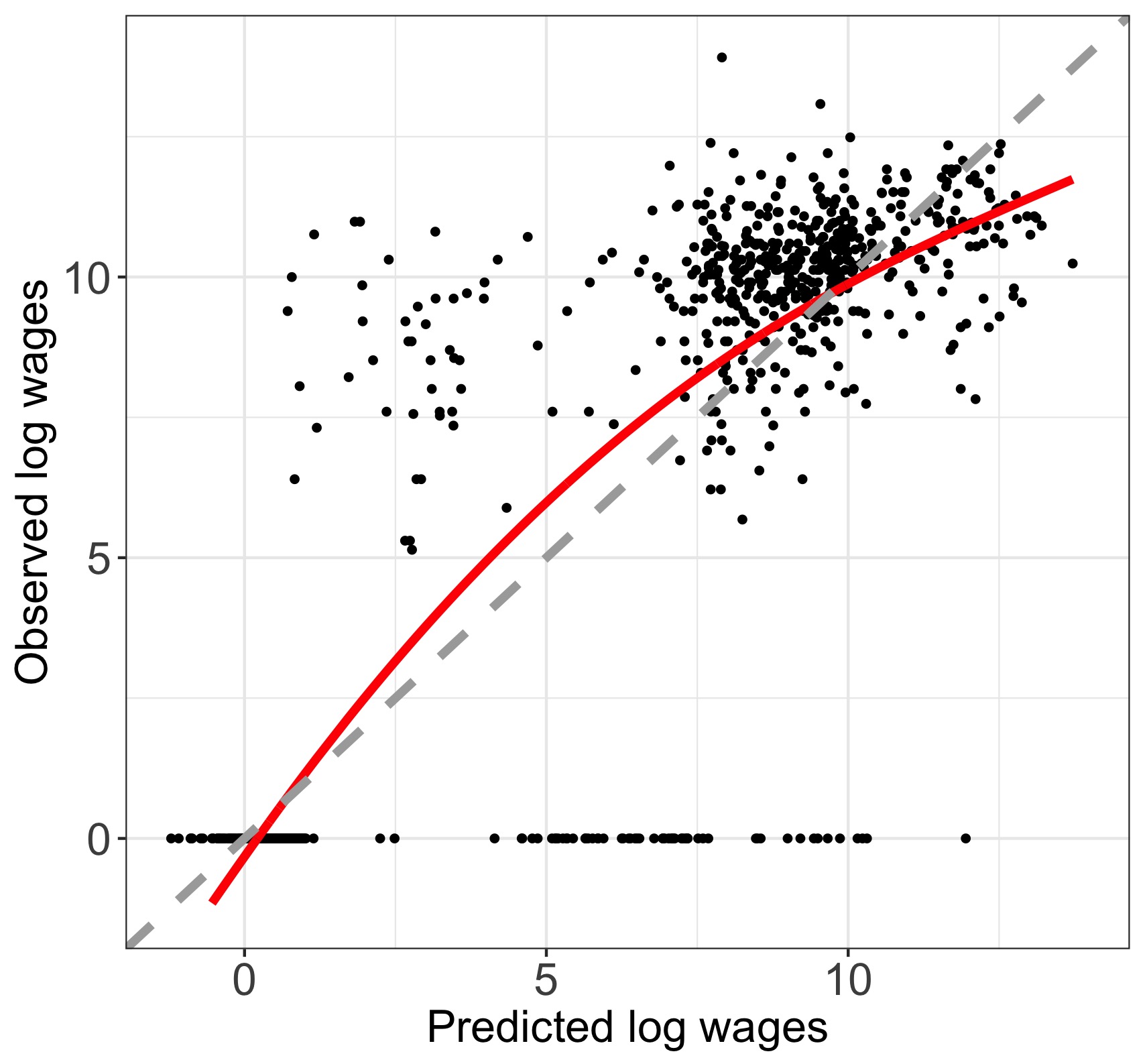} &
\includegraphics[width=0.4\textwidth]{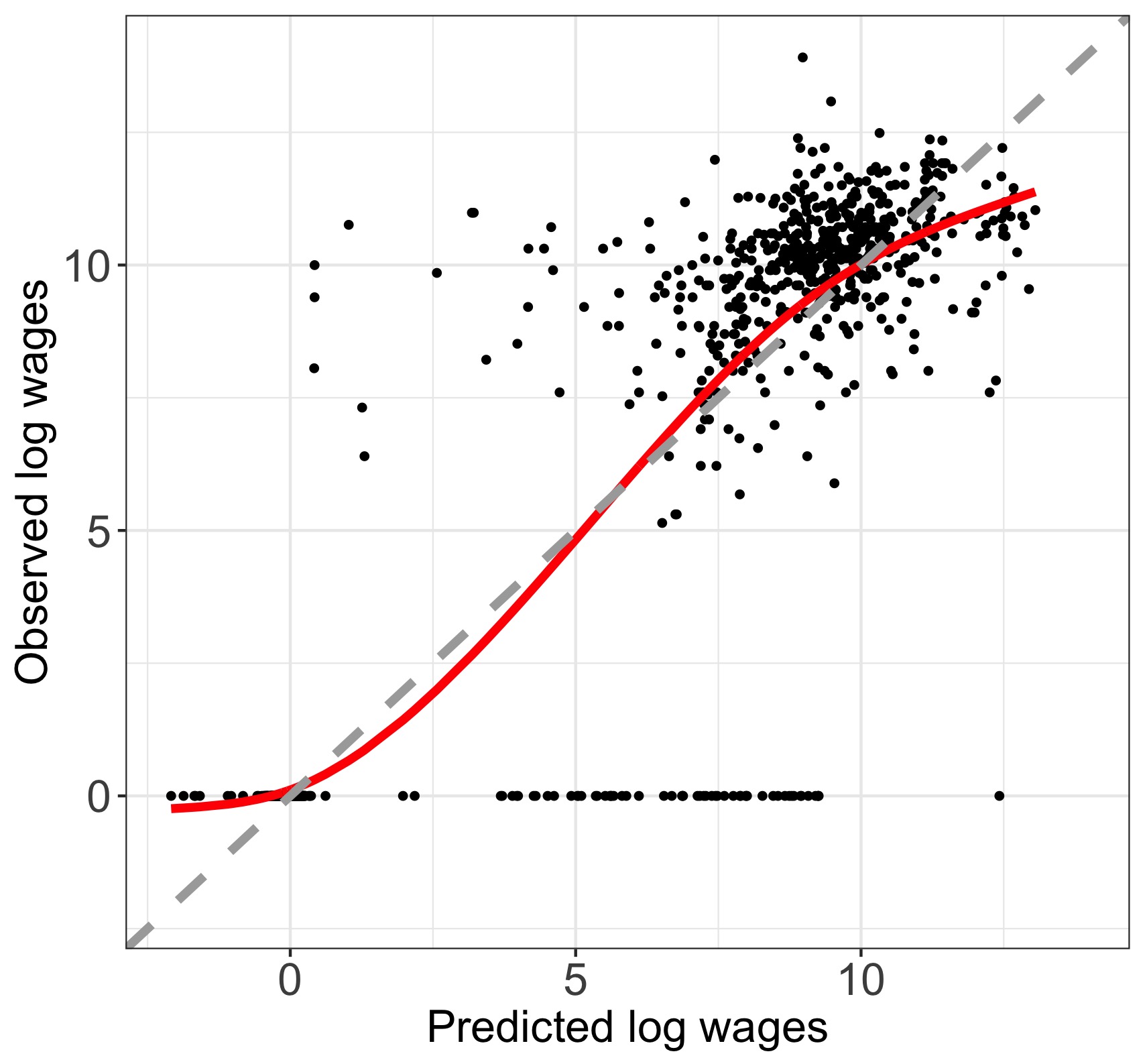} \\
Least squares & Lasso with interactions \\
\includegraphics[width=0.4\textwidth]{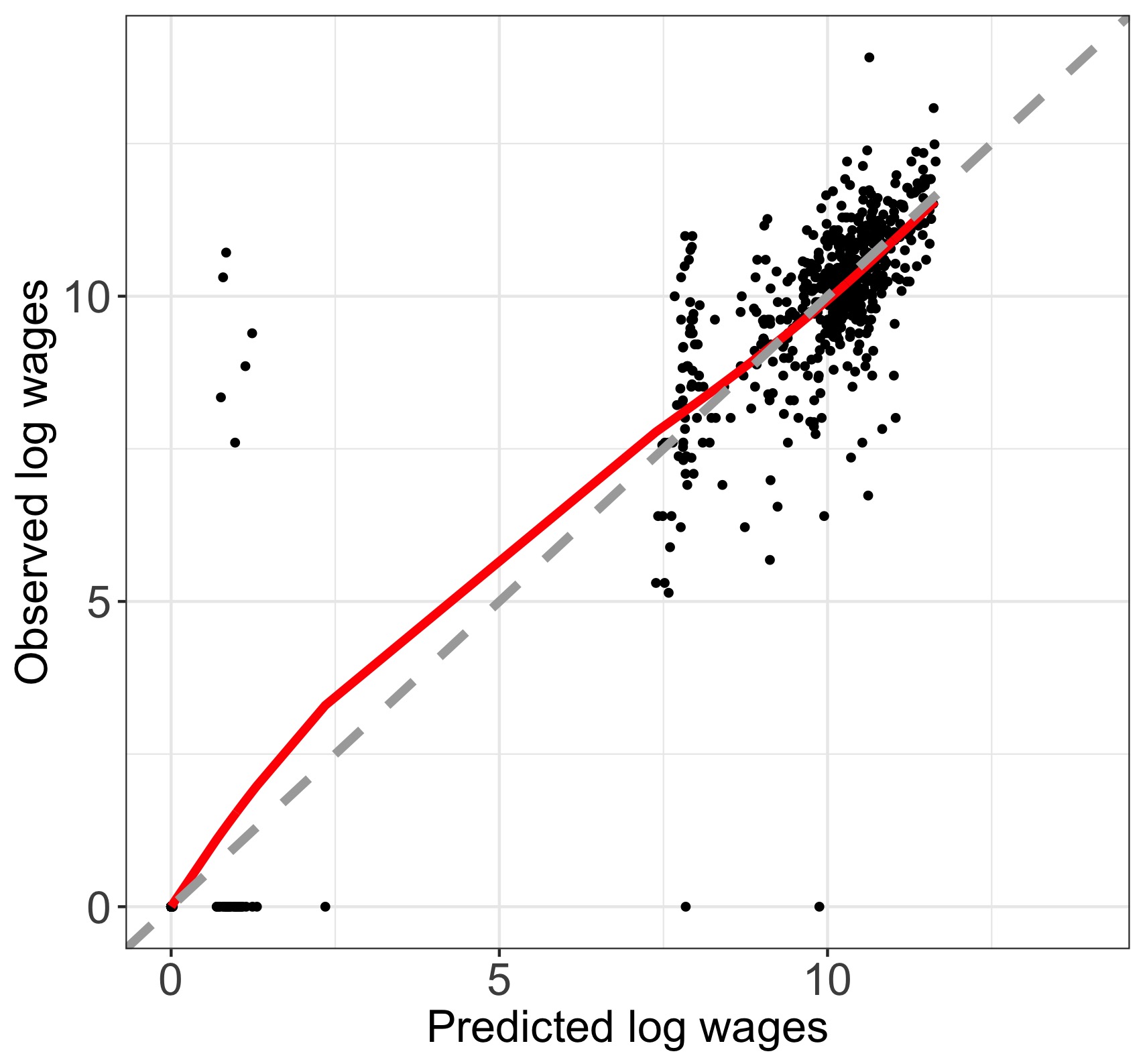} &
\includegraphics[width=0.4\textwidth]{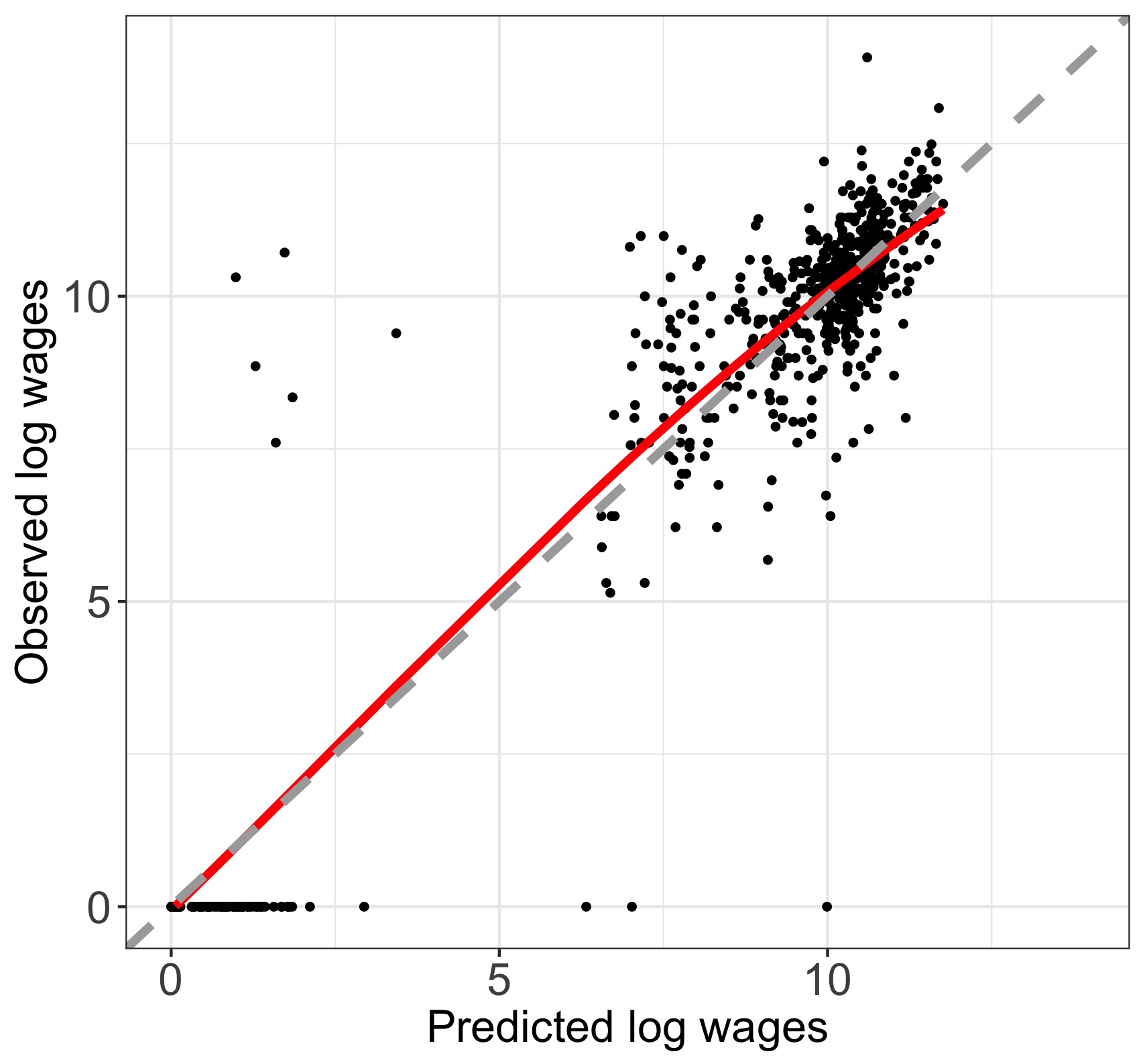} \\
Random forest & Local linear forest \\
\end{tabular}
\caption{Observed log wages versus model predictions for test set observations, based on (left to right, top to bottom) ordinary least squares, lasso with interaction terms, random forests, and local linear forests. These algorithms were trained on 10,000 points and evaluated on all remaining test points with families over 6 people. A cubic spline fit and the diagonal 45 degree line are also shown (as the full and dashed lines, respectively); the closer the spline fit is to the diagonal, the more evidence for calibration on this test set.}\label{fig-calibration}
\end{figure}

One motivation for studying wages is to compare wages across different types of workers, which requires accurate predictions even for types of workers who are less frequently observed. We thus evaluate predictive performance in several sparse regions of the covariate space, showing in Table \ref{wages-table} that local linear forests fit well in these regions.  To further explore this idea, 
 Figure \ref{fig-calibration} shows plots of observed log wages by predictions from ordinary least squares, lasso, random forests, and local linear forests, on individuals reporting a family size over 6, who amount for $3.3\%$ of the observed population. 
Cubic spline fits for each method are included to help evaluate calibration on this relatively sparse region of the dataset. 
Paired t-tests on the sets of squared errors for OLS ($t= 9.96$), lasso ($t=8.59$), boosting ($t = 2.92$), BART ($t = 3.27$), and random forests ($t=2.90$) give evidence for the improvements of local linear forests.

\subsection{Related Work}\label{sec-relatedwork}

Random forests were first introduced by \citet*{breiman2001}, building on the work of \citet*{breiman1984classification} on recursive
partitioning (CART), \citet*{Breiman:1996:BP:231986.231989} on bagging, and \citet*{Amit:1997:SQR:263023.263042}
on randomized trees.
\citet*{buhlmann2002analyzing} show how bagging makes forests smoother than single trees, while
\citet*{Biau:2012:ARF:2503308.2343682} and \citet*{scornet2015} establishes asymptotic risk consistency of
random forests under specific assumptions.
More sophisticated tree-based ensembles motivated by random forests have been proposed by
\citet*{Basu201711236}, who iteratively grow feature-weighted tree ensembles that perform especially well
for discovering interactions, \citet*{zhou2018boulevard}, who consider a hybrid between random
forests and boosting, and \citet*{zhu2015reinforcement}, who do deeper search during splitting
to mitigate the greediness of CART.  \citet*{2017arXiv170709461L} propose a Bayesian regression tree ensemble tailored to learning smooth, sparse signals and prove posterior minimaxity under certain conditions, highlighting the promise of tree-based methods that can adapt to smoothness.

The idea of considering random forests as an adaptive kernel method has been proposed by several papers. 
\citet*{hothorn} suggest using weights from survival trees and gives compelling simulation results,
albeit to our knowledge no theoretical guarantees. 
\citet*{meinshausen2006quantile} proposes this technique for quantile regression forests
and gives asymptotic consistency of the resulting predictions. 
\citet*{athey2016} leverage this idea to present generalized random forests as a method for
solving heterogeneous estimating equations. They derive an asymptotic distribution and confidence intervals for the resulting predictions. 
Local linear forests build on this literature; the difference being that we use the kernel-based
perspective on forests to exploit smoothness of $\mu(\cdot)$ rather than to target more complicated
estimands (such as a quantile).

Early versions of confidence intervals for random forests, backed by heuristic arguments and empirical
evidence, were proposed by \citet*{sexton2009standard} and \citet*{Wager:2014:CIR:2627435.2638587}.
\citet*{Mentch:2016:QUR:2946645.2946671} then established asymptotic normality of random forests
where each tree depends on a small subsample of training examples (so that there may be asymptotic bias), while \citet*{wager2017} provided a characterization of forests that allows for larger subsamples, deriving both asymptotic normality and valid confidence intervals.
The confidence intervals proposed here are motivated by the algorithm of \citet*{sexton2009standard},
and build on the random forest delta method developed by \citet*{athey2016}, taking advantage of improved subsampling rates for improved coverage.

As mentioned in the introduction, a predecessor to this work is a paper by \citet*{bloniarz2016supervised},
who consider local linear regression with supervised weighting functions, including
ones produced by a forest. The main differences between our method and that of \citet*{bloniarz2016supervised}
is that they do not adapt the tree-splitting procedure to account for the local linear
correction, and do not consider algorithmic features---such as ridge penalization---that appear
to be needed to achieve good performance both in theory and in practice. 
Additionally, our method is flexible to forests targeting any heterogeneous estimating equation, and in particular to causal forests. 
On the formal side, \citet*{bloniarz2016supervised} prove consistency of their method; however, they do not establish rates of convergence and thus, unlike in our Theorem \ref{main-result}, they cannot use smoothness
of $\mu(\cdot)$ to provide theoretical guarantees on improved convergence properties of the forest. They also do not provide a central limit theorem or confidence intervals.

More broadly, there is an extensive body of work on model-based trees that explores different
combinations of local regression and trees. \citet*{Torgo:1997:FMR:645526.657280} and
\citet*{Gama:2004:FT:990375.990395} study functional models for tree leaves,
fitting models instead of local averages at each node. \citet*{Karalic:1992:ELR:145448.146775}
suggests fitting a local linear regression in each leaf, and  \citet*{Torgo:1997:FMR:645526.657280} highlights the performance
of kernel methods in general for MOB tree methods. 
\citet{10.1007/978-3-642-23783-6_29} propose oblique random forests that learn split directions using the results from ridge regression, similar to our work developing splitting rules for local linear forests but more in the spirit of linear discriminant analysis (LDA). 
Case-specific random forests, introduced by \citet*{doi:10.1080/10618600.2014.983641}, use local information to upweight training samples not at the prediction step, but during the bootstrap to generate datasets for each tree. 
 \citet*{doi:10.1198/106186008X319331}, and
later  \citet*{doi:10.1080/00949655.2012.658804}, propose not only prediction, but recursive partitioning via fitting
a separate model in each leaf, similar to the residual splitting strategy of local linear forests.
Local linear forests complement this literature; they differ, however, in treating forests as a
kernel method. The leaf nodes in a local linear forest serve to provide neighbor information,
and not local predictions. 

Our work is motivated by the literature on local linear regression and maximum likelihood estimation 
\citep*{abadie2011bias,cleveland1988locally,GVK19282144X,heckman1998matching,
Load:1999,10.2307/3532868,Stone1977,tibshirani1987local}.
\citet*{Stone1977} introduces local linear regression and gives asymptotic consistency properties.
\citet*{Clev1979} expands on this by introducing robust locally weighted regression,
and \citet*{fan1992} give a variable bandwidth version.
\citet*{cleveland1988locally} explore further uses of locally weighted regression. 
Local linear regression has been particularly well-studied for longitudinal data, as in \citet*{li2010} and \citet*{yao2005}. 
\citet*{10.2307/2959068} use local polynomials to estimate the value of a function at the boundary of its domain.
\citet*{abadie2011bias} show how incorporating a local linear correction improves nearest neighbor matching procedures.

\section{Local Linear Forests}\label{sec-methodindetail}

Local linear forests use a random forest to generate weights that can then serve as a kernel for local linear regression. 
Suppose we have training data $\smash{(X_1, Y_1), \dots, (X_n, Y_n)}$ with $\smash{Y_i = \mu(X_i) + \e_i}$. 
Consider using a random forest to estimate the conditional mean function $\mu({x_0}) = \E[Y\mid X={x_0}]$ at a fixed test point ${x_0}$.
Traditionally, random forests are viewed as an ensemble method, where tree predictions are averaged to obtain the final estimate. Specifically, for each tree $T_b$ in a forest of $B$ trees, we find the leaf $L_b({x_0})$ with predicted response $\hat{\mu}_b({x_0})$, which is simply the average response of all training data points assigned to $L_b({x_0})$. We then predict the average $\smash{\hat{\mu}({x_0}) = (1/B) \sum_{b=1}^B \hat{\mu}_b({x_0})}$. 

An alternate angle, advocated by \citet*{hothorn}, \citet*{meinshausen2006quantile}, and \citet*{athey2016}, entails viewing random forests as adaptive weight generators. Equivalently write $\hat{\mu}({x_0})$ as 
\begin{align*}
\hat{\mu}({x_0}) 
&= \frac1B \sum_{b=1}^B \sum_{i=1}^n Y_i \frac{1\{X_i\in L_b({x_0})\}}{|L_b({x_0})|}  = \sum_{i=1}^n Y_i \frac1B \sum_{b=1}^B \frac{1\{X_i\in L_b({x_0})\}}{|L_b({x_0})|} = \sum_{i=1}^n \alpha_i({x_0}) Y_i,
\end{align*}
where the forest weight $\alpha_i({x_0})$ is 
\begin{equation}\label{eq-weights}
\alpha_i({x_0}) = \frac1B \sum_{b=1}^B \frac{1\{X_i\in L_b({x_0})\}}{|L_b({x_0})|}
\end{equation}
Notice that by construction, for each $i$, $0\le \alpha_i({x_0})\le 1$. Moreover, given that in at least one tree there exists a nonempty cell containing $x_0$, $\sum_{i=1}^n \alpha_i({x_0}) = 1$; otherwise all weights are equal to zero. 
\citet*{athey2016} use this perspective to harness random forests for solving weighted estimating equations, and give asymptotic guarantees on the resulting predictions. 

Local linear forests solve the locally weighted least squares problem \eqref{loss} with weights \eqref{eq-weights}.
Equation (\ref{loss}) has a closed-form solution, given below, following the closed-form solutions for ridge regression and classical local linear regression.
Throughout this paper, we let $A$ be the diagonal matrix with $A_{i,i} = \alpha_i({x_0})$, and let $J$ denote the $d+1\times d+1$ 
diagonal matrix with $J_{1,1} = 0$ and $J_{i+1,i+1} = 1$, so as to not penalize the intercept. 
We define $\Delta$, the centered regression matrix with intercept, as
$\Delta_{i,1} = 1$ and $\Delta_{i,j+1} = x_{i,j} - x_{0,j}$. 
Then the local linear forest estimator can be explicitly written as
\begin{equation}\label{estimator}
\begin{pmatrix} \hat{\mu}({x_0}) \\ \hat{\theta}({x_0}) \end{pmatrix} = \left(\Delta^T A \Delta + \lambda J \right)^{-1} \Delta^T A Y.
\end{equation} 
Define $\gamma_i = e_i \left(\Delta^T A \Delta + \lambda J \right)^{-1} \Delta^T$, where $e_i$ is a vector of zeroes with 1 in the $i$-th column.
Qualitatively, we can think of local linear regression as a weighting estimator, with $\gamma_i \alpha_i(x_0)$ a modulated weighting function whose ${x_0}$-moments are better aligned with the test point ${x_0}$:
\smash{$\hat{\mu}({x_0}) = \sum_{i = 1}^n \gamma_i \alpha_i({x_0}) Y_i$} with
\smash{$\sum_{i = 1}^n \gamma_i \alpha_i({x_0}) = 1$} and
\smash{$\sum_{i = 1}^n \gamma_i \alpha_i({x_0}) X_i\approx {x_0}$}, where
the last relation would be exact without a ridge penalty (i.e., with $\lambda = 0$).

With the perspective of generating a kernel for local linear regression in mind, we move to discuss the appropriate splitting rule for local linear forests. 

\subsection{Splitting for Local Regression}\label{sub-sec-splits}

Random forests traditionally use Classification and Regression Trees (CART) from \citet*{breiman1984classification} splits, which proceed as follows. 
We consider a parent node $P$ with $n_P$ observations $(x_{1}, Y_{1}), \dots, (x_{n_P}, Y_{n_P})$. 
For each candidate pair of child nodes $C_1, C_2$, we take the mean value of $Y$ inside each child, $\bar{Y}_1$ and $\bar{Y}_2$.
Then we choose $C_1, C_2$ to minimize the sum of squared errors
\begin{equation*}
\sum_{i: X_i\in C_1} (Y_i - \bar{Y}_i)^2 + \sum_{i: X_i\in C_2} (Y_i - \bar{Y}_2)^2. 
\end{equation*}
Knowing that we will use the forest weights to perform a local regression, we neither need nor want to use the forest to model strong, smooth signals; the final regression step can model them. 
Instead, in the parent node $P$, we run a ridge regression to predict $Y_{i}$ from $X_{i}$:
\begin{align}\label{splitting-rule}
\hat{Y}_{i} = \hat{\alpha}_P + x_{i}^T \hat{\beta}_P,
\end{align}
for intercepts $\hat{\alpha}_P$ and $\hat{\beta}_P = (x_P^T x_P + \lambda J)^{-1} x_P^T Y_P.$
We then run a standard CART split on the residuals $\smash{Y_{i} - \hat{Y}_{i}}$, modeling local effects in the forest and regressing global effects back in at prediction. 
Observe that, much like the CART splitting rule, an appropriate software package can enforce that a forest using this splitting rule splits on every variable and gives balanced splits; hence this splitting rule may be used to grow honest and regular trees (Section \ref{sec-theory}).

To explore the effects of CART and residual splitting rules, we consider this simulation first introduced by \citet*{friedman1991}.
Generate $X_1, \dots, X_n$ independently and identically distributed $U[0,1]^5$ and model $Y_i$ from  
\begin{equation}\label{friedman} y = 10 \sin(\pi X_{i1} X_{i2}) + 20(X_{i3} - 0.5)^2 + 10 X_{i4} + 5 X_{i5} + \epsilon,
\end{equation}
for $\e \sim N(0,\sigma^2)$. 
This model has become a popular study for evaluating nonparametric regression methods; see for example \citet*{chipman2010} and \citet*{TaddyBF}. 
It is a natural setup to test how well an algorithm handles interactions $  \sin(\pi X_{i1} X_{i2})$, its ability to pick up a quadratic signal $20(X_{i3} - 0.5)^2$, and how it simultaneously models strong linear signals $10 X_{i4} + 5 X_{i5} $. 

\begin{figure}[!t]
\begin{center}
\begin{tabular}{c}
\includegraphics[width=0.49\textwidth]{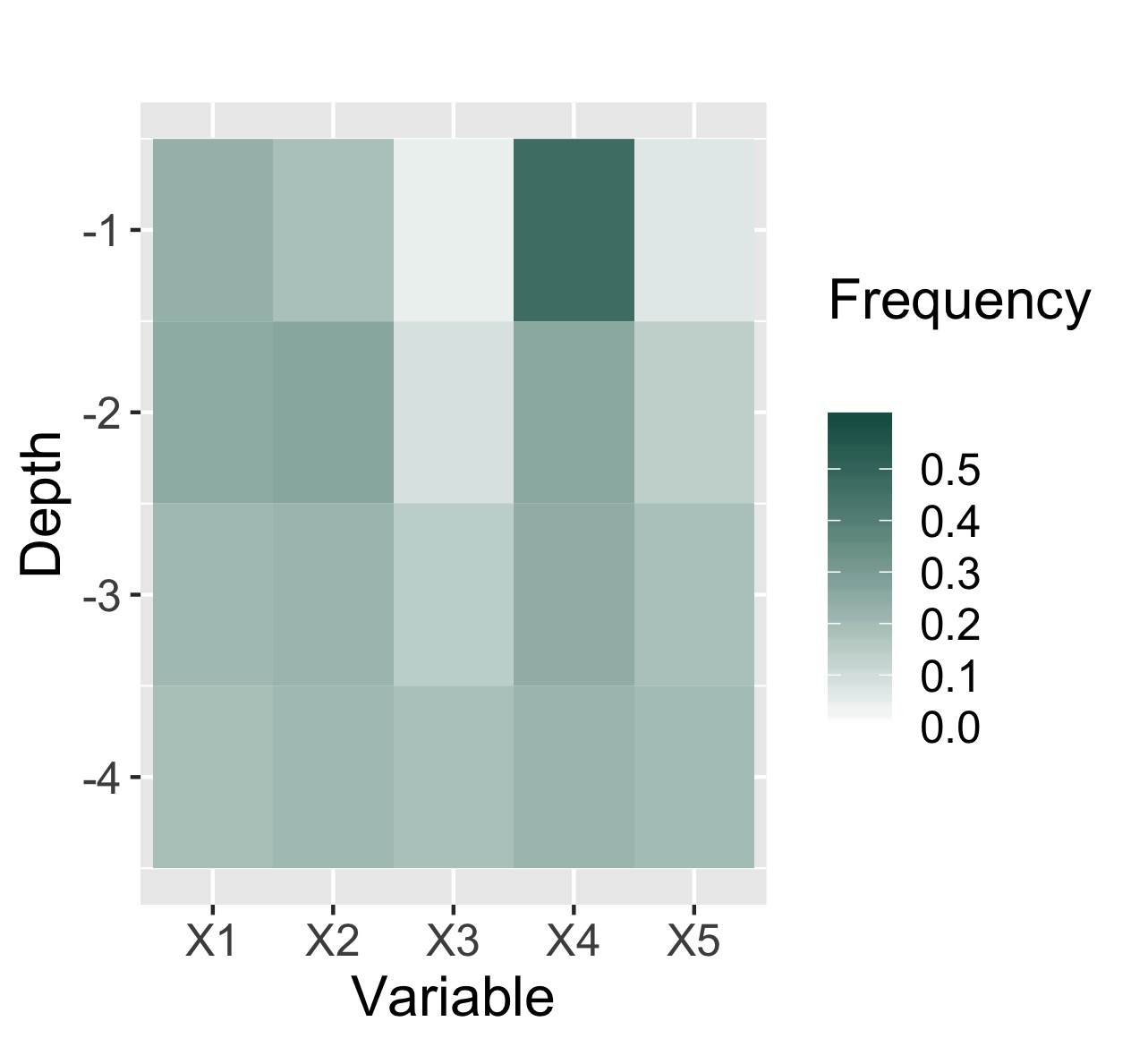} \\
CART split frequencies \\
\includegraphics[width=0.49\textwidth]{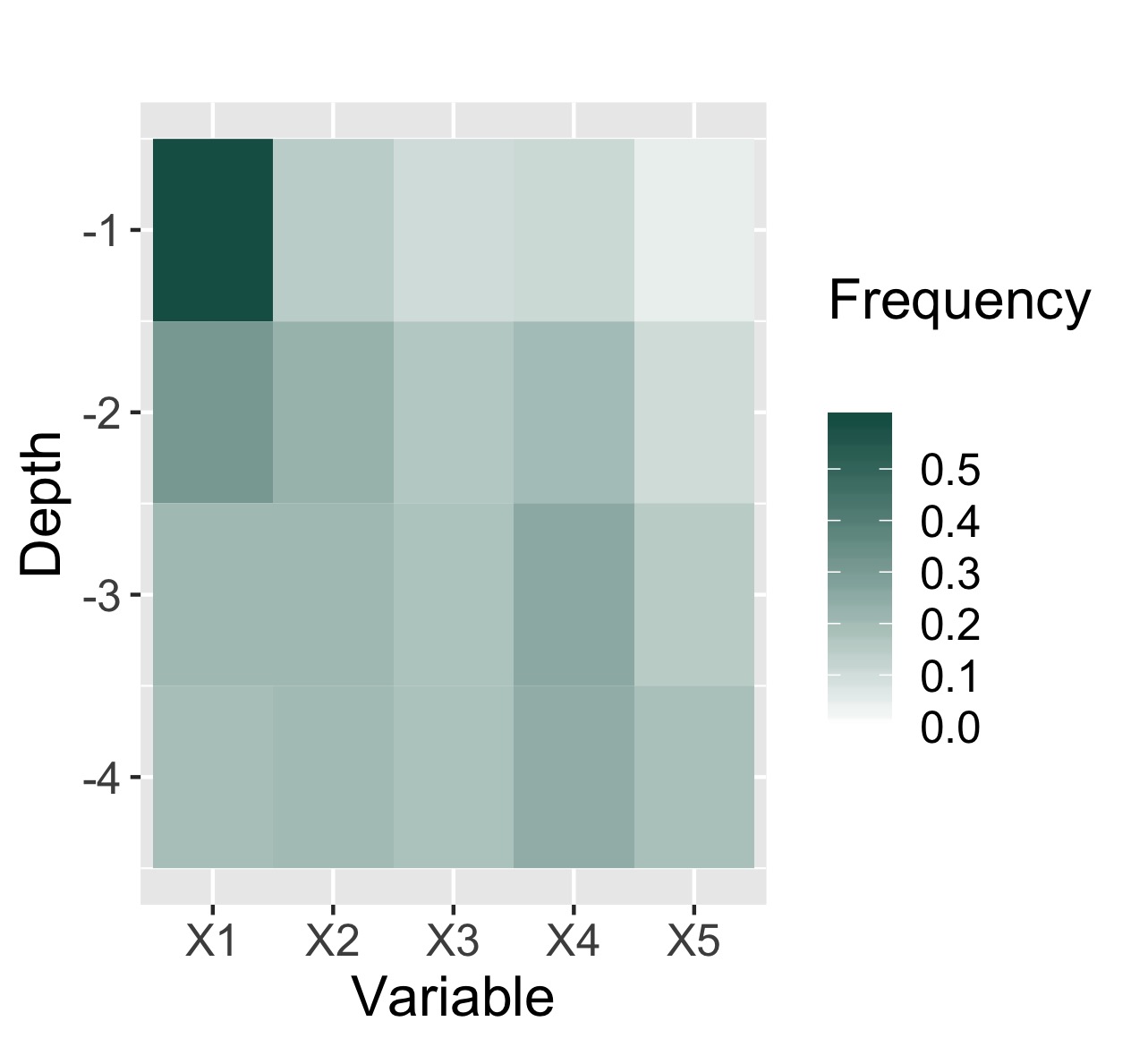}  \\
LLF split frequencies  \\
\end{tabular}
\caption{Split frequency plot for CART splits from an honest random forest (left) and residual splits from a local linear forest (right). Each forest was trained on $n=600$ observations from the data-generating process in \ref{friedman}. Variables 1 through 5 are on the x-axis, and the y-axis gives tree depth, starting with depth 1 at the top of the plot. Variables on which the forest splits frequently at depth $j$ have a dark tile in row $j$.}\label{friedman-figure-method}
\end{center}
\end{figure}

Figure \ref{friedman-figure-method} displays the split frequencies from an honest random forest (left) using standard CART splits, and a local linear forest (right). 
The x-axis is indexed by variable, here 1 through 5, and the y-axis gives tree depth for the first 4 levels of tree splits. 
Tiles are darkened according to how often trees in the forest split on that variable; a darker tile denotes more splits at that tree depth.
CART splits very frequently on variable 4, which contributes the strongest linear signal, especially at the top of the tree but consistently throughout levels. 
Local linear forests rarely split on either of the strong linear signals, instead spending splits on the three that are more difficult to model. 

\subsection{The Value of Local Linear Splitting}

\begin{figure}[!t]
\begin{center}
\includegraphics[width=0.49\textwidth]{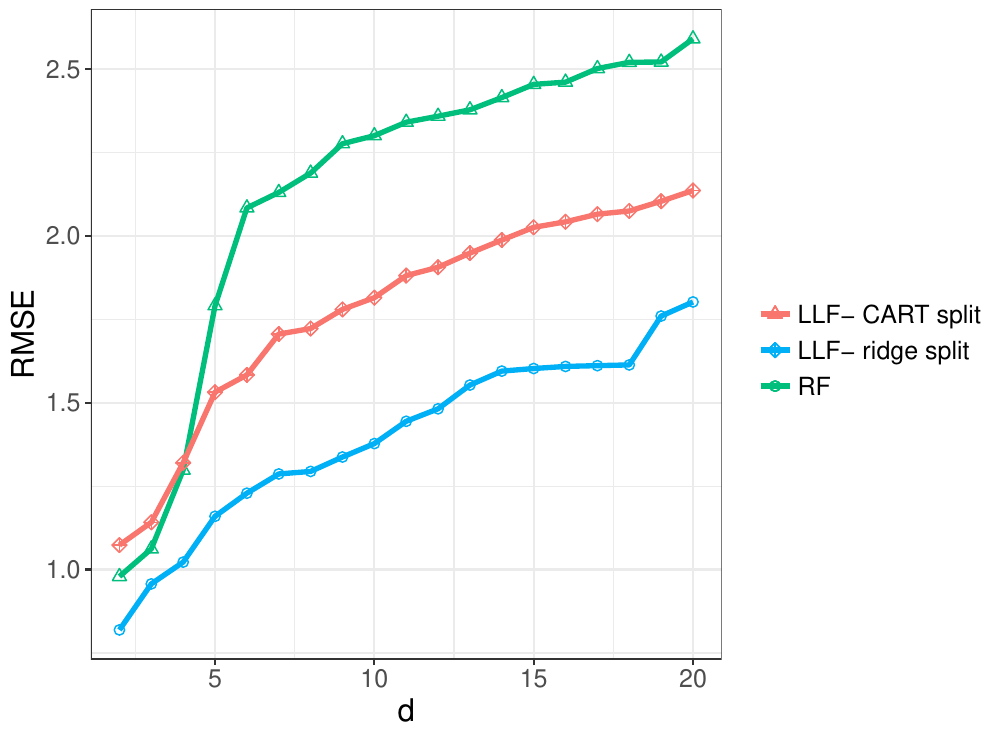}
\caption{Results from testing different splitting rules on data generated from equation \ref{split-eq}. Here the x-axis is dimension $d$, varying from 2 to 20, and we plot the Root Mean Square Error of prediction from random forests and from local linear forests with CART splits and with the ridge residual splits. We let $n=600$ and check results on $600$ test points at $50$ runs for each value of $d$.}\label{split-image-mean square error}
\end{center}
\end{figure}

Consider the following experiment, which highlights the benefit of the proposed splitting rule. 
We generate $X_1, \dots, X_n$ independently and uniformly over $[0,1]^{d}$. 
We hold a cubic signal $20(X_{i1} - 0.5)^3$ constant across simulations, and on each run  increase the dimension and add another linear signal. Formally, we let $\xi_j = \one\{j \le d\}$ and generate responses 
\begin{equation}\label{split-eq}
y_i = 20(X_{i1} - 0.5)^3\xi_1  +  \sum_{j=2}^3 10 X_{ij} \xi_j + \sum_{j=4}^5 5 X_{ij} \xi_j + \sum_{j=6}^{20}  2 X_{ij} \xi_j.
\end{equation}
For example, at simulation $3$  we have $\xi_1, \xi_2, \xi_3 = 1$ and hence we model $y_i = 20(X_{i1} - 0.5)^3 + 10 X_{i2} +10 X_{i3}$. 
Root Mean Square Error is displayed in Figure \ref{split-image-mean square error}.

In low dimension and with few linear signals, all three methods are comparable.
However, they begin to differ quickly.
Random forests are not designed for models with so many global linear signals, and hence their Root Mean Square Error increases dramatically with $d$. Moreover, as we add more linear effects, the gap between the two candidate splitting rules grows; heuristically, it becomes more important not to waste splits, and the residual splitting rule gives greater improvements. 
At a certain point, however, the gap between splitting rules stays constant. 
Once the forests simply cannot fit a more complex linear function with a fixed amount of data, the marginal benefits of the residual splitting rule level out. 
We show this to emphasize the contexts in which this splitting rule meaningfully affects the results. 

\subsection{Honest Forests}\label{sec-honesty}

Unless noted otherwise, all random forests used in this paper are grown using a type of sub-sample
splitting called ``honesty'', used by \citet*{wager2017} to derive the asymptotic distribution of random forest prediction.
As outlined in Procedure 1 of \citet*{wager2017}, each tree in an honest forest is grown using two non-overlapping subsamples
of the training data, denoted $\mathcal{I}_b$ and $\mathcal{J}_b$. We first choose a tree structure $T_b$ using only the data in 
$\mathcal{J}_b$, and write ${x_0} \leftrightarrow_{b} x'$ as the boolean indicator for whether the points ${x_0} $ and $ x'$ fall into
the same leaf of $T_b$. Then, in a second step, we define the set of neighbors of ${x_0}$ as
$L_b({x_0} ) = \{i \in \mathcal{I}_b : {x_0} \leftrightarrow_{b} X_i\}$; this neighborhood function is what we then use
to define the forest weights in \eqref{eq-weights}. We do not use the observed outcomes $y$ from sample $\mathcal{I}_b$ to select split points; but, to ensure that each node has a certain fraction of observations from its parent, we may use the covariates from $\mathcal{I}_b$. This modification allows us to grow honest forests that comply with the upcoming assumption \ref{regularity}, which says that trees are symmetric in permutations of training data index, split on every variable with nonzero probability, and balance parent observations in each child node. In this way, our theory is consistent and matches the implementation available online.

This type of subsample-splitting lets us control for potential overfitting when growing the tree $T_b$, because the
samples $\mathcal{J}_b$ which are in the neighborhood $L_b({x_0})$ were held out when growing $T_b$.
Despite considerable interest in the literature, there are no available consistency results for
random forests with fully grown trees that do not use honesty. \citet*{Biau:2012:ARF:2503308.2343682} uses a different type of sample splitting, wherein for each tree the data is split into two sets ($\mathcal{D}_n$ and $\mathcal{D}_n'$). $\mathcal{D}_n'$ is used to evaluate the CART criterion at each node during tree growth, and $\mathcal{D}_n$ is used to split. \citet*{biau2008consistency} and \citet*{wager2015uniform} rely on large leaves, while the results of \citet*{scornet2015} on fully grown trees rely on an unchecked high-level assumption. All of these choices come at a cost; forests grown to smaller leaves can model meaningful signal while averaging out erroneous splits. 
We build honest forests by default. 

Empirically, honesty can improve or worsen predictions.
In particular, with small samples sizes and strong signals, honesty may limit the expressive power of forests and thus
hurt predictive performance; conversely, with large sample sizes and weak signals, honesty may stabilize forests and
thus improve performance (see Appendix B of \citet{wager2017} for a discussion).
In any case, local linear corrections can help mitigate the loss of expressive power due to honesty, and so we may expect that
requiring honesty is less onerous with local linear forests than with regression forests.

\subsection{Tuning a Local Linear Forest}

We recommend selecting ridge penalties by cross-validation, which can be done automatically in the R package \texttt{grf}. 
It is often reasonable to choose different values of $\lambda$ for forest training and for local linear prediction. 
During forest growth, equation (\ref{splitting-rule}) gives ridge regression predictions $x_i^T \hat{{\bf \beta}}_P$ in each parent leaf. 
As trees are grown on subsamples of data, over-regularization at this step is a danger even in large leaves. 
Consequently, small values of $\lambda$ are advisable for penalization on regressions during forest training. 
Furthermore, as we move to small leaves, computing meaningful regression coefficients becomes more difficult; the ridge regression can begin to mask signal instead of uncovering it. 
A heuristic that performs well in practice is to store the regression estimates $\hat{{\bf \beta}}_P$ on parent leaves $P$. 
When the child leaf size shrinks below a cutoff, we use $\hat{{\bf \beta}}_P$ from the parent node to calculate ridge residual pseudo-outcomes, instead of estimating them from unstable regression coefficients on the small child dataset. 
In practice, this helps to avoid the pitfalls of over-regularizing and of regressing on a very small dataset when growing the forest. 
At the final regression prediction step (\ref{estimator}), however, a larger ridge penalty can control the variance and better accommodate noisy data.

With increasingly high-dimensional data, feature selection before prediction can significantly reduce error and decrease computation time. 
Often, a dataset will contain only a small number of features with strong global signals.
In other cases, a researcher will know in advance which variables are strong predictors that should be included in the linear smoothing, or of special interest. 
In these cases, it is reasonable to run the regression prediction step on this smaller subset of predictors expected to contribute overarching trends.
Such covariates, if they are not already known, can be chosen by a stepwise regression or lasso, or any other technique for automatic feature selection.
Last, it is worth noting that these tuning suggestions are pragmatic in nature; the theoretical guarantees provided in Section \ref{sec-theory} are for local linear forests trained without these heuristics.

\section{Extension to Causal Forests}\label{sec-causal}

For conciseness, the majority of this paper focuses on local linear forests for non-parametric regression; however, a similar local linear correction can also be applied to quantile regression
forests \citep*{meinshausen2006quantile}, causal forests \citep*{wager2017} or, more broadly, to any instance of generalized
random forests \citep*{athey2016}. To highlight this potential, we detail the method and discuss an example for heterogeneous
treatment effect estimation using local linear causal forests.

As in \citet*{athey2016}, we frame our discussion in terms of the Neyman-Rubin causal model \citep*{imbens2015causal}.
Suppose we have data $(X_i, Y_i, W_i)$, where $X_i$ are covariates, $Y_i \in \mathbb{R}$ is the response, and $W_i \in \{0, \, 1\}$ is the treatment.
In order to define the causal effect of the treatment $W_i$, we posit potential outcomes for individual $i$, $Y_i(0)$ and $Y_i(1)$,
corresponding to the response the subject would have experienced in the control and treated conditions respectively; we
then observe $Y_i = W_iY_i(1) + (1 - W_i)Y_i(0)$. We seek to estimate the conditional average treatment effect (CATE) of $W$,
namely $\tau(x) = \E[Y(1) - Y(0) \mid X=x]$. Throughout this paper we assume uncounfoundedness \citep*{doi:10.1093/biomet/70.1.41},
\begin{equation}
\left \{Y_i(0), \, Y_i(1)\right\} \indep W_i \, | \, X_i,
\end{equation}
and overlap, $\mathbb{P}[W_i = w] > 0$ for all $i$ and for each $w \in \{0,\, 1\}$. 
\citet*{wager2017} proposed an extension of random forests for estimating CATEs, and \citet*{athey2016} improved on the
method by making it locally robust to confounding using the transformation of \citet*{robinson1988root}.
Here, we propose a local linear correction to the method of \citet*{athey2016}, the orthogonalized causal forest, to strengthen
its performance when $\tau(\cdot)$ is smooth.

Local linear causal forests start as orthogonalized causal forests do, by estimating the nuisance components
\begin{equation}
\label{eq:nuisance}
e({x_0}) = \mathbb{P}[W_i = 1 \, |\,X_i = {x_0}] \text{ and } m({x_0}) = \mathbb{E}[Y_i \, |\,X_i = {x_0}]
\end{equation}
using a local linear forest. 
We then estimate the conditional average treatment effect via
\begin{equation}
\label{eq:CFll}
\begin{split}
&\left\{\hat{\tau}({x_0}), \, \hat{\theta}_\tau({x_0}), \, \hat{a}({x_0}), \, \hat{\theta}_a({x_0}) \right\} = \argmin_{\tau, \, \theta} \bigg\{ \sum_{i = 1}^n \alpha_i({x_0}) \Big( Y_i - \hat{m}^{(-i)}(X_i) - a - (X_i-{x_0}) \theta_a \\
&\ \ \ \ \ \ \ \ \ \ - \left( \tau  + \theta_\tau (X_i- {x_0}) \right) \left( W_i - \hat{e}^{(-i)}(X_i) \right) \Big)^2\ + \lambda_\tau \left\lVert \theta_\tau \right\rVert_2^2 
+ \lambda_a \left\lVert \theta_a \right\rVert_2^2 \bigg\},
\end{split}
\end{equation}
where the $^{(-i)}$-superscript denotes leave-one-out predictions from the nuisance models. If nuisance estimates are accurate, the intercept $\hat{a}$ should be 0; however, we leave it in the optimization for robustness.
We cross-validate local linear causal forests to select $\lambda_\tau$ and $\lambda_a$ by minimizing
the $R$-learning criterion recommended by \citet*{nie2017learning}:
\begin{equation}
\widehat{\text{Err}}\left(\hat{\tau}(\cdot)\right) = \sum_{i = 1}^n \left(Y_i - \hat{m}^{(-i)}(X_i) - \hat{\tau}(X_i) \left(W_i - \hat{e}^{(-i)}(X_i)\right)\right)^2.
\end{equation}
Observe that, analogous to the regression case, for very large values of $\lambda_a$ and $\lambda_{\tau}$, we will recover estimates from a causal forest. From the perspective, we can see local linear causal forests as an adapted R-learner method; note that \citet*{nie2017learning} and \citet*{kennedy2020optimal} give quasi-oracle and double-robustness properties of R-learner estimates. 

\subsection{Empirical Example: Attitudes to Welfare}
To illustrate the value of the local linear causal forests, we consider a popular dataset from the General Social Survey (GSS)
that explores how word choice reveals public opinions about welfare \citep{welfare}. 
Individuals filling out the survey from 1986 to 2010 answered whether they believe the government spends too much, too little, or the right amount on the social safety net. 
GSS randomly assigned the wording of this question, such that the social safety net was either described
as ``welfare'' or ``assistance to the poor". 
This change had a well-documented effect on responses due to the negative perception many Americans have about welfare; 
moreover, there is evidence of heterogeneity in the CATE surface \citep*{doi:10.1093/poq/nfs036}.

\begin{figure}[!t]
\centering
\begin{tabular}{cc}
\includegraphics[width=0.49\textwidth]{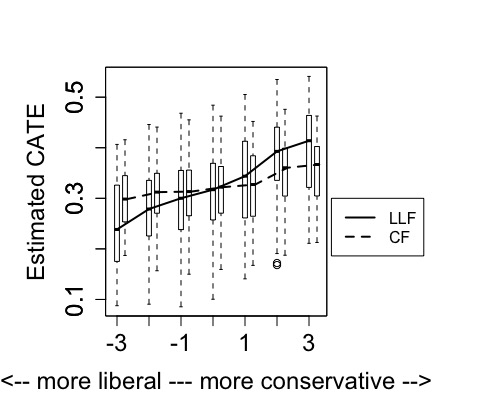} &
\includegraphics[width=0.49\textwidth]{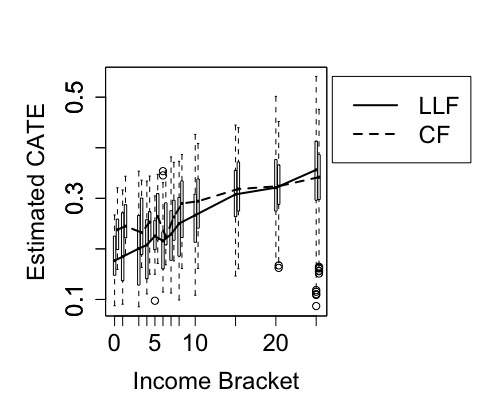} \\
CATE predictions by political views & CATE predictions by income \\
\end{tabular}
\caption{Trends in CATE predictions on the effect of the word ``welfare'' on people's perceptions of the social safety net. The left panel shows boxplots of CATE predictions among categories of political views, with dashed lines and straight lines connecting the medians of the causal forest and local linear causal forest predictions, respectively. The right panel shows analogous results for categories of income. All predictions are from cross-validated forests trained on 1000 training points and evaluated on 2000 test points.}\label{fig-welfare}
\end{figure}

Here, we write $W_i = 1$ if the $i$-th sample received the ``welfare'' treatment, and define $Y_i = 1$ if the $i$-th response was that the government spends too much on the social safety net.
Thus, a positive treatment effect $\tau(x)$ indicates that, conditionally on $X_i= x$, using the phrase ``welfare'' as
opposed to ``assistance to the poor'' increases the likelihood that the $i$-th subject says the government spends too much on the social safety net.
We base our analysis on $d = 12$ covariates, including income, political views, age, and number of children.
The full dataset has $N = 28,646$ observations; here, to make the problem interesting, and in particular relevant for practitioners who often have more limited survey data, we test our method on smaller subsamples of the data. Figure \ref{fig-welfare} shows boxplots of CATE predictions by category of political views and income, comparing local linear causal forests and causal forests, and indicating possible heterogeneity with an approximately linear pattern. 

In order to compare the performance of both methods, we use the transformed outcome metric of
\citet*{athey2015machine}. Noting that $\E[(2W_i - 1) Y_i \mid X_i] = \tau(X_i)$, they suggest examining the following test set
error criterion
\begin{equation}
\label{tau-error}
\begin{split}
&\mathcal{E} = \frac{1}{| \mathcal{S}_{test} |} \sum_{i \in \mathcal{S}_{test}} \left((2W_i - 1)Y_i - \hat{\tau}(X_i)\right)^2, \\
&\E[\mathcal{E}] = \E\left[\left(\tau(X) - \hat{\tau}(X)\right)^2\right] + S_0, \ \ 
S_0 = \E\left[\left((2W_i - 1)Y_i - \tau(X_i)\right)^2\right].
\end{split}
\end{equation}
If we can estimate $S_0$ and subtract it out, then \eqref{tau-error} gives an unbiased estimate of the mean-squared error of $\hat{\tau}(\cdot)$.
Here, we estimate $S_0$ via out-of-bag estimation on the full dataset with $N = 28,646$, assuming that a local linear forest
with such a large sample size has negligible error.

\begin{table}[t]
\begin{center}
\begin{tabular}{@{}llllllll@{}}
\toprule \toprule
Subsample size && 200 & 400 & 800 & 1200 & 1500 & 2000 \\
\midrule
Causal forest && 0.035 & 0.021 & 0.015 & 0.014 & 0.011 & 0.007 \\
Local linear causal forest && 0.027 & 0.017 & 0.013 & 0.013 & 0.011 & 0.006 \\
\bottomrule
\bottomrule 
\end{tabular}
\end{center}
\caption{Estimated in-sample mean square error \eqref{tau-error} of estimating the treatment effect on subsampled welfare data, averaged over 200 runs at each subsample size. We show estimated error from local linear causal forests and standard causal forests.
Tuning parameters were selected via cross-validation using the R-learner objective.}\label{table-welfare}
\end{table}

Table \ref{table-welfare} has error estimates for both types of forests using \eqref{tau-error}, and
verifies that using the local linear correction improves empirical performance across different
subsample sizes. Practically, we can view this change as enabling us to get good predictions on less data, a powerful improvement in cases like survey sampling where data can be expensive and difficult to attain. 
Section \ref{sec-simulations} contains a more detailed simulation study of local linear causal forests, comparing them with a wider array of baseline methods.

\section{Asymptotic Theory}\label{sec-theory}

Returning to the regression case, before we delve into the main result and its proof, we briefly discuss why the asymptotic behavior of local linear forests cannot be directly derived from the existing results of \citet*{athey2016}. This is due to a key difference in the dependence structure of the forest. 
In the regression case, a random forest prediction at ${x_0}$ is 
\smash{$\hat{\mu}_{\text{rf}}({x_0}) = \sum_{i=1}^n \alpha_i({x_0}) Y_i$},
where, due to honesty, $Y_i$ is independent of $\alpha_i({x_0})$ given $X_i$.
This conditional independence plays a key role in the argument of \citet*{wager2017}.
Analogously to \smash{$\hat{\mu}_{\text{rf}}({x_0})$}, we can write the local linear forest prediction as a weighted sum,
\begin{align}
\label{m-lambda-def}
\hat{\mu}({x_0}) = \sum_{i=1}^n \alpha_i({x_0}) \rho_i, \ \
\rho_i = e_1^T M_{\lambda}^{-1} \begin{pmatrix} 1 \\ X_i- {x_0} \end{pmatrix} Y_i, \ \
M_{\lambda} = \Delta^TA\Delta + \lambda J,
\end{align}
where we use notation $\Delta, \, A, \, J$ from \eqref{estimator}.
At a first glance, $\hat{\mu}({x_0})$ indeed looks like the output of a regression forest trained on observations $\rho_i$.
However, 
the dependence structure of this object is different. 
In a random forest, we make $Y_i$ and $\alpha_i({x_0})$ independent by conditioning on $X_i$. 
For a local linear forest, however, conditioning on $X_i$ will not guarantee that  $\rho_i$ and $\alpha_i({x_0})$ are independent,
thus breaking a key component in the argument of \citet*{wager2017}. 

\subsection{Main Result}

We now give a Central Limit Theorem for local linear forest predictions, beginning by stating
assumptions on the forest following those made in \citet*{wager2017}.

\begin{assumption}\label{regularity}(Regular Trees)
We assume that the forest grows regular trees: that the trees are symmetric in permutations of training data index, split on every variable with probability bounded from below by some probability $\pi>0$, and the trees are grown to depth $k$ for some $k \in \mathbb{N}$; and the trees and are balanced in that each split puts at least a fraction $\omega>0$ of parent observations into each child node. 
\end{assumption}

\begin{assumption}\label{honesty}(Honest Forests) 
We assume that the forest is honest as described in Section \ref{sec-honesty}, meaning that two distinct and independent subsamples are selected for each tree. Only the outcome values from one subsample are used to select the splits, and only those from the other to estimate parameters in the nodes. 
\end{assumption}

Finally, in our proof, we use the following high-level assumption on the distribution of samples as
weighted by the random forest kernel. Let $d_\alpha = \sum_{i = 1}^n \alpha_i(x_0) (X_i - x_0)$ denote
the difference between $x_0$ and the $\alpha_i$-weighted average of $X_i$, and let
$S_\alpha = \sum_{i = 1}^n \alpha_i (X_i - x_0)^{\otimes 2}$ be the associated quadratic form. From
Jensen's inequality, we immediately see that $d_\alpha' S_\alpha^{-1} d_\alpha \leq 1$, with equality only in
the degenerate case where $X_i$ has no variation in the direction of $d_\alpha$. The following assumption rules out
such degenerate distributions of $X_i$ within leaves, and requires that the $X_i$ enough variation
along $d_\alpha$ to separate $d_\alpha' S_\alpha^{-1} d_\alpha$ from 1. We believe this to be a reasonable assumption that
should be satisfied by any reasonable tree-growing algorithm; and it would be of interest to derive it from first principles in future work.

\begin{assumption}\label{leaf}(Leaf Distribution) 
We train our forest such that $1 - d_\alpha' S_\alpha^{-1} d_\alpha = \Omega\p{1}$.
\end{assumption}

Subsampling plays a central role in our asymptotic theory, allowing us to prove asymptotic normality
by building on the work of \citet*{efron1981jackknife}.
Moreover, subsampling is what we use to tune the bias-variance trade-off of the forest:
Forests whose trees are grown on small subsamples have lower bias but higher variance (and vice-versa).

In order to establish asymptotic unbiasedness of forests, \citet*{wager2017} require a subsample size of
at least $n^{\beta}$, with 
\begin{equation} \beta_{\text{rf}} = 1 - \left( 1 +  \frac{d}{\pi} \frac{\log(\omega)}{\log(1-\omega)}\right)^{-1} < \beta < 1. \label{beta-rf}
\end{equation}
This convergence rate of a traditional honest random forest, however, does not improve when $\mu({x_0})$ is smooth.
Here, we show that by using a local regression adjustment and assuming smoothness of $\mu(\cdot)$,
we can grow trees on smaller subsamples of size (\ref{beta-llf}) without sacrificing asymptotic variance control.
This allows us to decrease the bias (and improve the accuracy) of our estimates.

Our main result establishes asymptotic normality of local linear forest predictions, and gives this improved subsampling rate.
The condition $\omega \le 0.2$ allows us to leverage theory from \citet*{wager2017} when controlling part of $\hat{\mu}(x_0)$. 
We prove this result in Appendix B.

\begin{theorem}\label{main-result}
Suppose that we have training data $Z_i = (X_i, Y_i)$ identically and independently distributed on $[0,1]^d \times \R$,
that $X_i$ has a uniform distribution on $[0,1]^d$, and let $x_0$ be a point in the interior of $[0,1]^d$.
Suppose furthermore that $\mu({x}) = \mathbb{E}[Y\mid X={x}]$ is differentiable with a Lipschitz continuous derivatives
$\mu_2({x}) = \mathbb{E}[Y^2\mid X={x}]$ is Lipschitz continuous,
that $\Var[Y\mid X={x_0}] > 0$, and that
$\E[|Y - \E[Y \mid X=x]|^{2+\delta} \mid X=x] \le M$ for some constants $M, \delta > 0$ over all $x \in [0,1]^d$.
Given this data-generating process, we consider local linear forests based on trees grown according to 
Assumptions \ref{regularity}, \ref{honesty} and \ref{leaf}, with $\omega \le 0.2$ and subsamples of size $s$ with $s = n^{\beta}$, for
\begin{equation}\label{beta-llf}
\beta_{\min} = 1 - \left( 1 +  \frac{d}{1.3 \pi} \frac{\log(\omega)}{\log(1-\omega)}\right)^{-1} < \beta < 1.
\end{equation}
We also use a ridge regularization parameter in \eqref{loss} that grows at rate
\begin{equation}
\lambda = \Theta \left( s^{-0.99 \frac{\log(1-\omega)}{\log(\omega)} \frac{\pi}{d}} \sqrt[4]{\frac{s}{n}} \right)
\label{eq-lambda-rate}
\end{equation}
Then, there is a sequence $\sigma_n({x_0}) \to 0$ such that 
\begin{equation*}
\frac{\hat{\mu}_n({x_0}) - \mu({x_0})}{\sigma_n({x_0})} \Rightarrow N(0,1), ~~  \sigma_n^2({x_0}) = \widetilde{O}\p{n^{-(1-\beta)}},
\end{equation*}
where $\widetilde{O}(\cdot)$ is a version of big-$O$ notation that ignores log-factors.
\end{theorem}

The main draw of this results is that the best attainable rate of convergence $\beta_{\min}$ is improved compared to the
rate $\beta_{\text{rf}}$ from \eqref{beta-rf} obtained in \citet*{wager2017}. The reason we were able to obtain such an improvement
in the rate of convergence is that we have assumed and consequently exploited smoothness via local linear regression.

We note that the condition \eqref{beta-llf} on the subsampling rate enforces undersmoothing, i.e., that the error of 
\smash{$\hat{\mu}_n({x_0})$} will be dominated by variance. Undersmoothing implies that our estimator is asymptotically
unbiased, and facilitates construction of confidence intervals. One limitation of this strategy is that it is in general difficult to tune
an algorithm for undersmoothing; in particular, tuning via cross-validation does not guarantee undersmoothing.
Developing methods for random forest inference that do not rely on undersmoothing following, e.g., \citet*{Hall_2013},
would be of considerable interest; however, this falls beyond the scope of the present paper.

\subsection{Pointwise Confidence Intervals}

This section complements our main result, as the Central Limit Theorem becomes far more useful when we have valid standard error estimates.
Following \citet*{athey2016}, we use the random forest delta method to develop pointwise confidence intervals
for local linear forest predictions.

The random forest delta method starts from a solution to a local estimating equation
with random forest weights $\alpha_i({x_0})$:
\begin{equation}\label{general-est-eqn}
\sum_{i = 1}^n \alpha_i({x_0}) \psi\left(X_i, \, Y_i; \, \hat{\mu}({x_0}), \, \hat{\theta}({x_0})\right) = 0.
\end{equation}
\citet*{athey2016} then propose estimating the error of these estimates as
\begin{equation}\label{delta-method}
\widehat{\Var}\left[\left(\hat{\mu}({x_0}), \hat{\theta}({x_0})\right)\right] = \widehat{V}({x_0})^{-1} \widehat{H}_n({x_0}) \left (\widehat{V}({x_0})^{-1}\right)',
\end{equation}
where $V({x_0}) = \nabla_{(\mu,\theta)} \E[\psi({x_0},Y;\mu,\theta)\mid {x_0}={x_0}]$ is the slope of the expected
estimating equation at the optimum, and \smash{$\widehat{H}_n({x_0})$} is an estimate of
\begin{equation}
\label{eq:Hn}
H_n({x_0}) = \Var\left[\sum_{i = 1}^n \alpha_i({x_0}) \psi\left(X_i, \, Y_i; \, \mu^*({x_0}), \, \theta^*({x_0})\right) \right].
\end{equation}
The upshot is that \smash{$H_n({x_0})$} measures the variance of an (infeasible) regression forest
with response depending on the score function $\psi$ at the optimal parameter values, and that we can in
fact estimate \smash{$H_n({x_0})$} using tools originally developed for variance estimation with
regression forests. Meanwhile, $V({x_0})$ can be estimated directly using standard methods.

With local linear forests, $(\hat{\mu}, \hat{\theta})$ solve \eqref{general-est-eqn} with score function
\begin{equation}\label{eq:psi}
\psi(Y_i,X_i;\mu,\theta) = \nabla_{(\mu, \, \theta)} \ \frac{1}{2} \left( \left( Y_i - \Delta_i \begin{pmatrix} \mu \\ \theta \end{pmatrix} \right)^2 + \lambda \left\lVert \theta \right\rVert_2^2 \right),
\end{equation}
where we again use notation defined in \eqref{estimator} and \eqref{m-lambda-def}. First, we note
that we have access to a simple and explicit estimator for $V({x_0})$: Twice differentiating \eqref{loss} with
respect to the parameters $(\mu, \, \theta)$ gives 
\begin{equation}\label{second-derivative}
\nabla^2_{(\mu, \, \theta)} \ \frac{1}{2} \left( \sum_{i=1}^n \alpha_i(x_0) (Y_i - \mu - \Delta_i {\theta})^2 + \lambda ||{\theta} ||_2^2 \right) = \sum_{i=1}^n \alpha_i({x_0}) \Delta_i^T \Delta_i + \lambda J = M_{\lambda},
\end{equation}
which we can directly read off of the forest. In this paper, we are only interested in confidence
intervals for $\mu({x_0})$, i.e., the first coordinate of $(\mu, \, \theta)$, and to estimate its variance
we only need access to the entry in the upper-left corner of \eqref{delta-method}, which we call \smash{$\hat{\sigma}_n^2$}.
Given our setting, we then note that we can re-express the relevant part of \eqref{delta-method} as follows, in terms of
$\zeta' = e_1'M_\lambda^{-1}$:
\begin{equation}
\begin{split}
&\hat{\sigma}_n^2 = \zeta' \widehat{H}_n(x) \zeta
=  \widehat{\Var}\left[\sum_{i = 1}^n \alpha_i({x_0}) \  \Gamma_i(\mu^*({x_0}), \, \theta^*({x_0}))\right], \\
&\Gamma_i(\mu, \, \theta) = \left(\zeta \cdot \Delta_i\right) \left( Y_i - \Delta_i \begin{pmatrix} \mu \\ \theta \end{pmatrix} \right),
\end{split}
\end{equation}
where \smash{$\widehat{\Var}[]$} refers to an estimate of the variance of the infeasible regression forest defined
between the brackets.

Next, we follow \citet*{athey2016}, and proceed using the bootstrap of little bags construction of
\citet*{sexton2009standard} to estimate the variance of this infeasible regression forest.
At a high level this method is a computationally efficient half-sampling
estimator. For any half sample $\mathcal{H}$, let $\Psi_{\mathcal{H}}$ be the average of the empirical
scores $\Gamma_i$ averaged over trees that only use data from the half-sample $\mathcal{H}$:
\begin{equation}
\Psi_{\mathcal{H}} = \frac{1}{\left|\mathcal{S}_{\mathcal{H}} \right|} \sum_{b \in \mathcal{S}_{\mathcal{H}}}
\frac{\sum_{i = 1}^n  1\left(\left\{X_i\in L_b({x_0})\right\}\right) \Gamma_i\left(\hat{\mu}({x_0}), \, \hat{\theta}({x_0})\right)}{\sum_{i = 1}^n  1\left(\left\{X_i\in L_b({x_0})\right\}\right)},
\end{equation}
where \smash{$\mathcal{S}_{\mathcal{H}}$} is the set of trees that only use data from the half-sample $\mathcal{H}$,
and $L_b({x_0})$ contains neighbors of ${x_0}$ in the $b$-th tree (throughout, we assume that the subsample used to grow
each tree has less than $n/2$ samples). Then, a standard half-sampling estimator would simply use \citep*{efron1982jackknife}
\begin{equation}
\label{eq:HS}
\hat{\sigma}_n^2 = \binom{n}{\lfloor n/2 \rfloor}^{-1}
\!\!\!\!\!
\sum_{\left\{\mathcal{H} \, : \, \left|\mathcal{H}\right| = \left\lfloor \frac{n}{2} \right\rfloor\right\}}
\left(\Psi_{\mathcal{H}} - \bar{\Psi}\right)^2, \ \
\bar{\Psi} = \binom{n}{\lfloor n/2 \rfloor}^{-1} \!\!\!\!\! \sum_{\left\{\mathcal{H} \, : \, \left|\mathcal{H}\right| = \left\lfloor \frac{n}{2} \right\rfloor\right\}} \Psi_{\mathcal{H}}.
\end{equation}
Now, carrying out the full computation in \eqref{eq:HS} is impractical, and naive Monte Carlo
approximations suffer from bias. However, as discussed in \citet*{athey2016} and
\citet*{sexton2009standard}, bias-corrected randomized algorithms are available and perform well.
Here, we do not discuss these Monte Carlo bias corrections, and instead refer to Section 4.1 of
\citet*{athey2016} for details.
Simulation results on empirical confidence interval performance are given in Section \ref{results}.

\section{Simulation Study}\label{sec-simulations}

\subsection{Methods}\label{cv}

In this section, we compare local linear forests, random forests, BART \citep*{chipman2010}, and gradient boosting \citep*{friedman2001}. 
We also include a lasso-random forest baseline for local linear forests: on half of the training data, run a lasso \citep*{Tibshirani94regressionshrinkage} regression; on the second half, use a random forest to model the corresponding residuals. 
Like local linear forests, this method combines regression and forests, making it a natural comparison; it is similar in spirit to the tree-augmented Cox model of \citet*{10.1093/biostatistics/kxi024}, who combine pruned CART trees with proportional hazards regression.

Random forests are trained using the R package \texttt{grf}  \citep{package-grf}, and are cross-validated via the 
default parameter tuning in \texttt{grf}, which selects values for mtry, minimum leaf size, sample fraction,
and two parameters (alpha and imbalance penalty) that control split balance. 
Local linear forests are tuned equivalently with additional cross-validation for regularization parameters. Variables for the regression at prediction are selected via the lasso.
Because existing theoretical results for random forests rely on honesty, all random forests are built with the honest construction.
All lasso models are implemented via \texttt{glmnet} \citep*{ref-glmnet-package} and cross-validated with their automatic cross-validation feature. 
Local linear regression is not included in these comparisons, since the implementations \texttt{loess} and \texttt{locfit} both fail for $d>6$ on this simulation; in Appendix A, we compare local linear regression with this set of methods on lower dimensional linear models. 
Unless otherwise specified, all reported errors are Root Mean Square Error on 1000 test points averaged over 50 simulation runs. 

Gradient boosted trees are implemented by the R package \texttt{XGBoost} \citep{xgboost_package}.
BART for treatment effect estimation is implemented following \citet*{hill2011bayesian}.
As is standard, we use the \texttt{BART} package \citep{bart_package} without any additional tuning.
The motivation for not tuning is that if we want to interpret the BART posterior in a Bayesian
sense (as is often done), then cross-validating on the prior is hard to justify; and in fact most existing
papers do not cross-validate BART.

\subsection{Simulation Design}

The first design we study is Friedman's example from equation (\ref{friedman}). 
Figure \ref{friedman-sims} shows errors at $n=1000$ fixed, with dimension $d$ varying from 10 to 50. There are two plots shown, to highlight the differences between error variance $\sigma=5$ and $\sigma=20$. 
Appendix A reports a grid of errors for dimensions 10, 30, and 50, with $n = 1000$ and $5000$, and $\sigma$ taking values of 5 and 20. 
The second design we consider is given in Section \ref{sec-introduction}, as in equation (\ref{boundary-eq}).
Again we test on a grid, letting dimension $d$ take values in 5 and 50, $n$ either 1000 or 5000, and $\sigma$ at 0.1, 1, and 2. Errors are reported in Appendix A.

The third simulation is designed to test how local linear forests perform in a more adversarial setting, where we expect random forests to outperform. We simulate $X_1, \dots, X_n$ i.i.d. $U[0,1]^d$ and model responses as 
\begin{equation}
\label{eq-step}
y_i = \frac{10}{1 + \exp(-10 * (X_{i1} - 0.5))} + \frac{5}{1 + \exp(-10 * (X_{i2} - 0.5))} + \e, ~~~ \e \sim N(0, 5^2).
\end{equation}
Here we test dimension $d = 5, 20$ and values of $n = 500, 2000, 10000$. For this simulation, we compare only honest random forests and local linear forests, in order to compare confidence intervals; we compute out of bag Root Mean Square Error and average confidence interval coverage and length. Results are reported in Table \ref{table-confidence}. 

\subsection{Results}\label{results}

\begin{figure}[!t]
\begin{center}
\begin{tabular}{cc}
\includegraphics[width=0.49\textwidth]{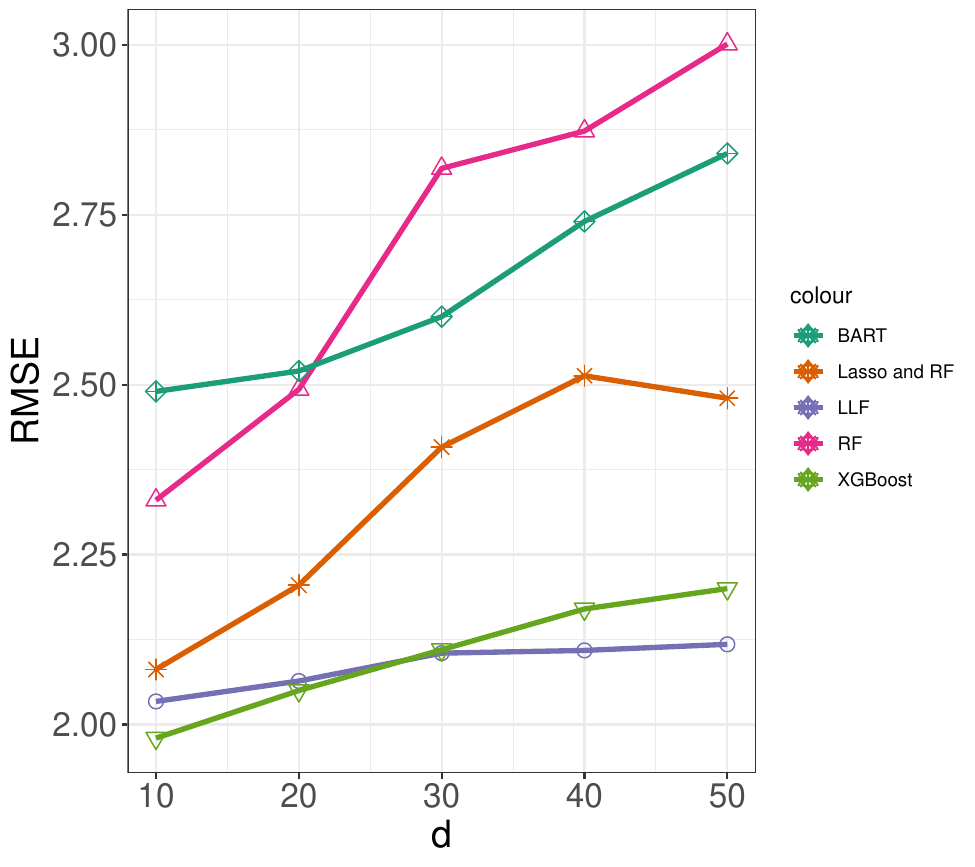} &
\includegraphics[width=0.49\textwidth]{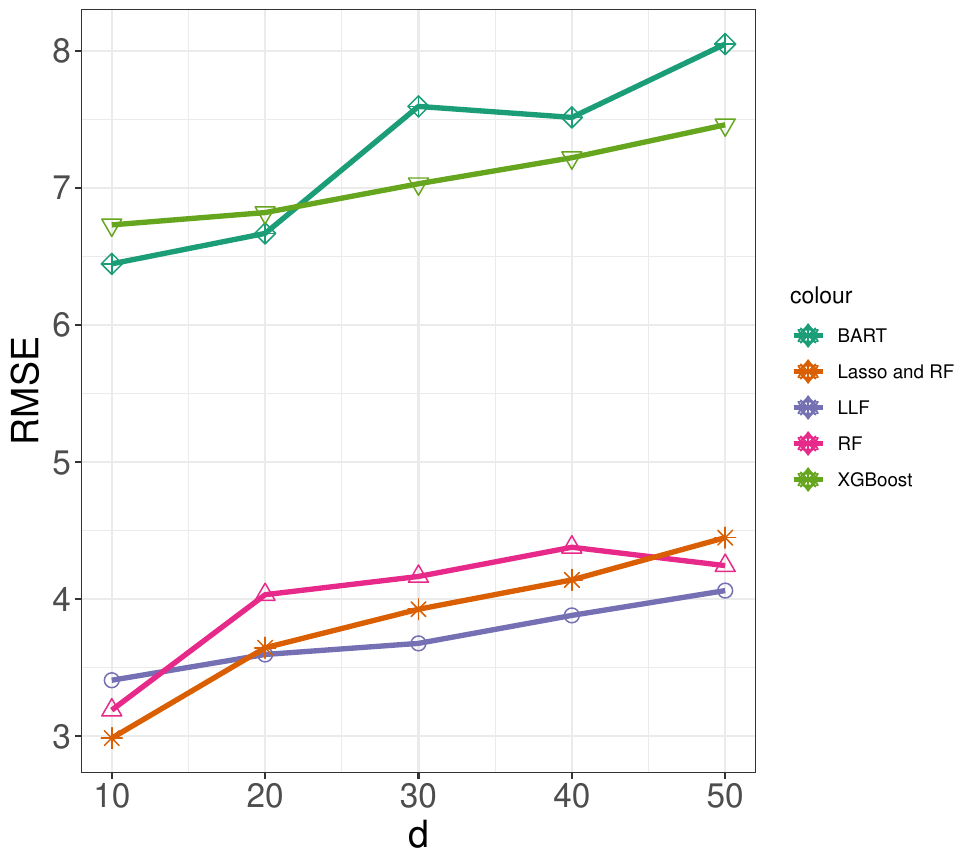} \\
Errors on low variance & Errors on high variance \\
\end{tabular}
\caption{Root Mean Square Error of predictions on $1000$ test samples from equation \ref{friedman}, with $n=1000$ held fixed and dimension $d$ varied from 10 to 50. Plots is shown for error standard deviation $\sigma=5$ (left) and $\sigma=20$ (right). Error was calculated in increments of 10, and averaged over 50 runs per method at each step. Methods evaluated are random forests (RF) local linear forests (LLF), lasso and random forests, boosted trees, and BART.}\label{friedman-sims}
\end{center}
\end{figure}

Figure \ref{friedman-sims} shows Root Mean Square Error from equation \ref{friedman} at $\sigma=5$ (left) and $\sigma=20$ (right).
In Section \ref{sec-methodindetail}, we showed that local linear forests and standard regression forests split on very different variables when generating weights. 
Our intuition is that these are splits we have saved; we model the strong linear effects at the end with the local regression, and use the forest splits to capture more nuanced local relationships for the weights.
Local linear forests consistently perform well as we vary the parameters, lending this credibility. 
The lasso-random forest baseline lines up closely with local linear forests in the high noise case, separating more for low noise. 
BART and random forests form the next tier of methods on the low noise case; in the high noise case, honest random forest are clustered with local linear forests. BART and boosting separate in the higher noise case, suffering compared to the other methods. 
Appendix A shows the fuller Root Mean Square Error comparison from Friedman's model.

We move to the second simulation setup, equation \ref{boundary-eq}, meant to evaluate how methods perform in cases with a strong linear trend in the mean. Tree-based methods will be prone to bias on this setup, as the forests cannot always split on $X_1$, and because the signal is global and smooth. 
Full error results on the range of competing methods are given in Appendix A.
Local linear forests do quite well here; they detect the strong linear signal in the tail, as we saw in Figure \ref{figure-confidence}, and model it successfully throughout the range of the feature space. Gradient boosted trees perform very well in the low-noise case, but their performance sharply declines when we increase $\sigma$. 

\begin{table}[t]
\begin{center}
\footnotesize
\begin{tabular}{@{}cccccccccccc@{}}
\toprule \toprule 
Setup & d & n && \multicolumn{2}{c}{Coverage} && \multicolumn{2}{c}{Length} && \multicolumn{2}{c}{Root Mean Square Error} \\
\midrule\midrule
Equation \ref{boundary-eq} & & && RF & LLF && RF & LLF && RF & LLF \\
\cmidrule(lr){5-6} \cmidrule(lr){8-9} \cmidrule(lr){11-12} 
& 5 & 500 && 0.90 & 0.94 && 2.40 & 2.35 && 0.63 & 0.55 \\
& 5 & 2000 && 0.97 & 0.96 && 2.23 & 1.85 && 0.43 & 0.35 \\ 
& 5 & 10000 && 0.97 & 0.98 && 1.41 & 2.20 && 0.28 & 0.42 \\
& 20 & 500 && 0.88 & 0.92 && 2.23 & 2.13 && 0.68 & 0.55 \\
& 20 & 2000 && 0.89 & 0.96 && 2.14 & 2.12 && 0.17 & 0.09 \\
& 20 & 10000 && 0.97 & 0.99 && 1.23 & 0.89  && 0.24 & 0.13 \\ 
\midrule\midrule
Equation \ref{friedman} & & && RF & LLF && RF & LLF && RF & LLF \\
\cmidrule(lr){5-6} \cmidrule(lr){8-9} \cmidrule(lr){11-12} 
& 5 & 500 && 0.54 & 0.65 && 3.56 & 3.82 && 2.36 & 2.03 \\
& 5 & 2000 && 0.63 & 0.69 && 3.17 & 3.21 && 1.77 & 1.58 \\
& 5 & 10000 && 0.70 & 0.75 && 2.75 & 2.77 && 1.32 & 1.18  \\ 
& 20 & 500 && 0.45 & 0.59 && 4.06 & 4.61 && 8.82 & 4.85 \\ 
& 20 & 2000 && 0.57 & 0.64 && 3.55 & 4.22 && 5.25 & 3.20 \\
& 20 & 10000 && 0.52 & 0.70 && 2.28 & 2.89 && 1.83 & 1.46  \\ 
\midrule
Equation \ref{eq-step} & & && RF & LLF && RF & LLF && RF & LLF \\
\cmidrule(lr){5-6} \cmidrule(lr){8-9} \cmidrule(lr){11-12} 
& 5 & 500 && 0.85 & 0.89 && 3.26 & 3.46 && 1.50 & 0.90 \\ 
& 5 & 2000 && 0.90 & 0.92 && 2.54 & 2.82 && 0.52 & 0.45 \\ 
& 5 & 10000 && 0.82 & 0.92 && 1.47 & 1.36 && 0.40 & 0.3 \\ 
& 20 & 500 && 0.85 & 0.89 && 3.36 & 3.19 && 1.50 & 0.98 \\
& 20 & 2000 && 0.90 & 0.90 && 2.71 & 2.36 && 0.62 &  0.46 \\
& 20 & 10000 && 0.87 & 0.92 && 1.94 & 1.55 && 0.58 & 0.37 \\ 
\bottomrule \bottomrule 
\end{tabular}
\end{center}
\caption{Average coverage and length of $95\%$ confidence intervals from honest random forests (RF) and local linear forests (LLF), along with Root Mean Square Error on the same out of bag (OOB) predictions. OOB coverage is averaged over 50 runs of the simulation setups in equations \ref{boundary-eq}, \ref{friedman}, and \ref{eq-step} and reported for the given values of dimension $d$ and number of training points $n$. We hold $\sigma=\sqrt{20}$ constant for equation \ref{boundary-eq}, and $\sigma=5$ constant for equation \ref{friedman}, and train on sample fraction $0.5$.}\label{table-confidence}
\end{table}

We also examine the behavior of our confidence intervals in each of the given simulation setups, shown here in Table \ref{table-confidence}. We give average coverage of $95\%$ confidence intervals from 50 repetitions of random forests and local linear forests on draws from the simulation setups in equations \ref{boundary-eq}, \ref{friedman}, and \ref{eq-step}, as well as average confidence interval length and Root Mean Square Error. On equation \ref{boundary-eq}, local linear forest confidence intervals are consistently shorter and closer to $95\%$ coverage, with correspondingly lower mean squared error. Here, both random forests and local linear forests achieve fairly low Root Mean Square Error and coverage at or above $88\%$. For the setup in equation \ref{friedman}, on the other hand, neither method achieves higher than $75\%$ coverage, and the local linear forest confidence intervals are longer than the random forest confidence intervals. This is an encouraging result, indicating that local linear forests confidence intervals are more adaptable to the context of the problem; we would hope for long confidence intervals when detection is difficult. Moreover, the poor coverage we see sometimes across both methods is likely because the confidence intervals are built on asymptotic results, which may not apply in some relatively low $n$ settings. 

We include the approximate step function in equation \ref{eq-step} to highlight a favorable example for random forests. Local linear forests see equivalent or better coverage on this setup, although at the cost of longer confidence intervals in low dimensions. Especially on small training datasets, local linear forests also improve on random forest predictions in Root Mean Square Error. 

The majority of these settings are well-suited to local linear forest success; one can think of several examples where the method is likely to under-perform. With a small number of covariates, the method is similar to local linear regression, possibly worse if the forest has overfit to the data. If the local linear correction does not accurately model any underlying smoothness, cross-validation will select large values for $\lambda$, but given speed and inaccurate assumptions, random forests would be preferred in this setting. 

\subsection{Local Linear Causal Forests}

In Section \ref{sec-causal}, we introduced a real-data example where the local linear extension of causal forests naturally applies. 
Evaluating errors empirically, however, is difficult, so we supplement that with a simulation also used by  \citet*{wager2017} in evaluating causal forests and \citet*{xlearner}, used to evaluate their meta-learner called the X-learner. 
Here we let $X \sim U([0,1]^d)$.
We fix the propensity $e(x) = 0.5$ and $\mu(x)=0$, and generate a causal effect $\tau$ from each
\begin{align}
\tau(X_i) = \zeta(X_{i1}) \zeta(X_{i2}), ~~~ \zeta(x) = \frac{2}{1+\exp(-20(x-1/3))} \label{sim1} \\
\tau(X_i) = \zeta(X_{i1}) \zeta(X_{i2}), ~~~ \zeta(x) = 1 + \frac{1}{1+\exp(-20(x-1/3))}. \label{sim2}
\end{align}
We will assume unconfoundedness \citep*{doi:10.1093/biomet/70.1.41}; therefore, because we hold propensity fixed, this is a randomized controlled trial. 

\begin{table}[t]
\begin{center}
\footnotesize
\begin{tabular}{@{}lllllllll@{}}
\toprule \toprule
&& \multicolumn{3}{c}{Simulation 1 (equation \ref{sim1})} & \phantom{a} & \multicolumn{3}{c}{Simulation 2 (equation \ref{sim2})} \\
\midrule
n && X-BART & CF & LLCF && X-BART & CF & LLCF  \\
\midrule
200 && 1.01 & 0.94 &{\bf 0.80} && {\bf 0.67} & 0.77 & 0.71   \\
400 && 0.76 & 0.50 & {\bf 0.47} && 0.56 & 0.55 & {\bf 0.50}  \\
600 && 0.61 & 0.39 & {\bf 0.35}  &&  0.50 & 0.41 & {\bf 0.38} \\
800 && 0.55 & 0.35 & {\bf 0.31}  &&  0.46 & 0.34 & {\bf 0.32}   \\ 
1000 && 0.50 & 0.33 & {\bf 0.30} && 0.44 & 0.29 & {\bf 0.28}  \\
1200 && 0.48 & 0.32 & {\bf 0.28}  && 0.42 & 0.27 & {\bf 0.26}  \\
\bottomrule
\bottomrule 
\end{tabular}
\end{center}
\caption{Average Root Mean Square Error of predicting the heterogeneous treatment effect $\tau_i$ on 100 repetitions of the simulation given in equation (\ref{sim1}). We vary the sample size $n$ from $200$ to $1200$ in increments of 200, always testing on $2000$ test points. We report errors from local linear causal forests (LLCF), causal forests (CF), and the X-learner with BART as base learner (X-BART). Minimizing errors are reported in bold.}\label{kunzel}
\end{table}

We compare local linear forests, causal forests, and X-BART, which is the X-learner using BART as a base-learner. 
Causal forests as implemented by \texttt{grf} are tuned via the automatic self-tuning feature.
As in the prediction simulation studies, we do not cross-validate X-BART because the authors recommend X-BART specifically for when a user does not want to carefully tune. We acknowledge that this may hinder its performance.
Local linear causal forests are tuned via cross-validation.
On these simulations, consider relatively small sample sizes ranging from $n=200$ to $n=1200$ with dimension $d=20$. The goal of this simulation is to evaluate how effectively we can learn a smooth heterogeneous treatment effect in the presence of many noise covariates. 
\citet*{wager2017} emphasize equation \ref{sim1} as a simulation that demonstrates how forests can suffer on the boundary of a feature space, because there is a spike near $x=0$. 
Root Mean Square Error over 100 repetitions is reported in Table \ref{kunzel}, demonstrating that local linear forests give a significant improvement over causal forests.
Both of these setups are reasonable tests for how a method can learn heterogeneity, and demonstrate potential for meaningful improvement with thoughtful variable selection and robustness to smooth heterogeneous signals. 

\section{Discussion} 

In this paper, we proposed local linear forests as a modification of random forests equipped to model smooth signals and fix boundary bias issues. 
We presented asymptotic theory showing that, if we can assume smoother signals, we can get better rates of convergence as compared to generalized random forests. 
We showed on the welfare dataset that local linear forests can model smooth heterogeneous causal effects, and illustrated when and why they outperform competing methods. 
We also gave confidence intervals from the delta method for the regression case, and demonstrated their effectiveness in simulations. 

The regression adjustments in local linear forests prove especially useful when some covariates have strong global effects with moderate curvature. Furthermore, the adjustment provides centered predictions, adjusting for errors due to an asymmetric set of neighbors. 
It may be that there is a useful polynomial basis corresponding to every situation where local linear forests performed well, but finding such a model would likely require hand-tuning the functional form for competitive performance, and is not automatically suited to a mix of smooth and non-smooth signals. For a departure from frequentist techniques, BART and Gaussian processes are both hierarchical Bayesian methods; BART can be viewed as a form of Gaussian process with a flexible prior, making BART the preferred baseline. 

There remains room for meaningful future work on this topic. In some applications, we may be interested in estimating the slope parameter $\theta({x_0})$, rather than merely accounting for it to improve the precision of $\mu({x_0})$. While local linear forests may be an appropriate method for doing so, we have not yet explored this topic and think it could be of significant interest. 
Extending our theoretical results beyond pointwise convergence would enable finding uniform confidence bands and be of considerable theoretical and practical interest. We have also not considered the theoretical or empirical improvements that could arise from assuming higher order smoothness in the functions we are estimating; searching for additional optimality results in this setting could be another interesting research question. 

\bibliographystyle{plainnat}
\bibliography{llf}  

\renewcommand{\thesubsection}{\Alph{subsection}}

\section*{Appendix}

\subsection{Remaining Simulation Results}

We include first Table \ref{friedman-table}, giving a full error comparison of the lasso-random forest baseline, BART, boosting, random forests, and local linear forests, on Friedman's data-generating process: generate $X_1, \dots, X_n$ i.i.d. $U[0,1]^5$ and model $Y_i$ from  
\begin{equation*} y = 10 \sin(\pi X_{i1} X_{i2}) + 20(X_{i3} - 0.5)^2 + 10 X_{i4} + 5 X_{i5} + \epsilon,
\end{equation*} 
Errors are reported on dimension ranging from 10 to 50, $\sigma$ from 5 to 20, and $n=1000$ and $5000$, averaged over 50 training runs. 

\begin{table}[h]
\begin{center}
\footnotesize
\begin{tabular}{@{}lllllllll@{}}
\toprule \toprule
$d$ & $n$ & $\sigma$ && RF  &lasso-RF & LLF & BART & XGBoost \\
10 & 1000 & 5 && 2.33 &2.12 & 2.03 & 2.49 & {\bf 1.98} \\
10 & 5000 & 5 && 1.90 &  {\bf 1.48} & 1.57 & 1.51 & 1.52 \\
30 & 1000 & 5 && 2.82 & 2.41 & {\bf 2.11} & 2.60 & {\bf 2.11} \\
30 & 5000  & 5 && 2.08 & {\bf 1.61} &1.73 & 2.03 & 1.64 \\
50 & 1000 & 5 && 3.00  & 2.48 &  {\bf 2.12} & 2.84 & 2.20 \\
50 & 5000 & 5 && 	2.18 &	1.82 & {\bf 1.80} &	2.11& 1.82 \\
\midrule\midrule
10 &1000	 &20	&& 	{\bf 3.19} & 3.41  & 3.40	&6.45 & 6.73 \\
10 & 5000	 & 20	&&	2.43	& 	2.35 & {\bf 2.29} &	3.85	& 4.42 \\
30 &1000	 & 20&&	4.17 &		 3.98 & {\bf 3.68} &	7.60	& 7.03 \\
30 &5000	 &20&&	2.97 &		2.66 	& {\bf 2.40} &4.78	 & 4.85 \\
50 &1000	 & 20	&&	4.25 &		4.45	& {\bf 3.88}	 &8.05&	7.47 \\
50 &5000	 & 20&&	3.16	&	2.67 & {\bf 2.35} &	4.95	& 4.97 \\
\bottomrule
\bottomrule 
\end{tabular}
\end{center}
\caption{Root mean  square error on Friedman's function, with dimension $d$ from 10 to 50 predictors in increments of 20, and consider error standard deviation $\sigma$ ranging from 1 to 20, for a variety of signal-to-noise ratios. For this setting, $\Var(\E[Y\mid X]) \approx 23.8$, as approximated over 10,000 Monte Carlo repetitions; so letting $\sigma=1$ corresponds to a signal-to-noise ratio of about 23.8, while letting $\sigma=20$ corresponds to a signal-to-noise ratio of about $0.24$. We train on $n=1000$ and $n=5000$ points, and report test errors from predicting on $1000$ test points. All errors reported are averaged over 50 runs and the methods are cross-validated as described in the main document. Minimizing errors are reported in bold.}\label{friedman-table}
\end{table}

We include next Table \ref{boundary-table}, again giving a more complete error comparison of the lasso-random forest baseline, BART, boosting, random forests, and local linear forests, on the data-generating process: simulate $X_1, \dots, X_n$ i.i.d. Uniform $[0,1]^{20}$, with responses  
\begin{equation*}
y_i = \log\left(1+\exp(6 X_{i1})\right) + \e, ~~ \e \sim \mathcal{N}(0, \, 20).
\end{equation*}
Errors are reported on dimension ranging from 5 to 20, $\sigma$ from 0.1 to 2, and $n=1000$ and $5000$, averaged over 50 training runs.

\begin{table}[h]
\begin{center}
\footnotesize
\begin{tabular}{@{}lllllllll@{}}
\toprule \toprule
$d$ & $n$ & $\sigma$ &&  RF & lasso- RF & LLF & BART & XGBoost \\
\midrule
5 & 1000 & 0.1 && 0.10  & 0.06 & {\bf 0.02} & 0.27 & 0.07 \\
5 & 5000 & 0.1 && 0.06 & {\bf 0.02} & {\bf 0.02} & 0.22 & 0.06 \\
50 & 1000 & 0.1 && 0.29  & 0.18 & 0.11 & 0.52 & {\bf 0.07} \\
50 & 5000 & 0.1 && 0.18  & 0.10 & 0.07 & 0.62 & {\bf 0.06} \\
\midrule
5 & 1000 & 1 && 0.21  & 0.24 & {\bf 0.14} & 0.47 & 0.56 \\
5 & 5000 & 1 && 0.15  & 0.11 & {\bf 0.09} & 0.26 & 0.52 \\
50 & 1000 & 1 && 0.41  & 0.39 & {\bf 0.20} & 0.82 & 0.53 \\
50 & 5000 & 1 && 0.23  & 0.21 & {\bf 0.10} & 0.57 & 0.52 \\
\midrule
5 & 1000 & 2 && 0.31 & 0.55 & {\bf 0.26} & 0.69 & 1.21 \\
5 & 5000 & 2 && 0.25  & 0.28 & {\bf 0.21} & 0.40 & 1.18 \\
50 & 1000 & 2 && 0.47  & 0.27 & {\bf 0.24} & 0.89 & 1.22  \\
50 & 5000 & 2 && 0.33  & 0.27 & {\bf 0.15} & 0.70 & 0.96 \\
\bottomrule
\bottomrule 
\end{tabular}
\end{center}
\caption{Root mean  square error from simulations on random forests, lasso-random forest, local linear forests, BART, and boosting. We vary sample size $n$, error variance $\sigma$, and ambient dimension $d$, and report test error on $1000$ test points. We estimate $\Var[\E[Y\mid X]]$ as $3.52$ over 10,000 Monte Carlo repetitions, so that signal-to-noise ratio ranges from $352$ at $\sigma=0.1$ to $0.88$ at $\sigma=2$. All errors are averaged over 50 runs, and minimizing errors are in bold.}\label{boundary-table}
\end{table}

To close this section, we consider some basic linear and polynomial models in low dimensions, in order to effectively compare local linear forests with local linear regression. We simulate $X \sim U[0,1]^3$ and model responses from two models, 
\begin{align}
y_i &= 10X_{i1} + 5X_{i12} + X_{i3} + \e \label{linear1}\\
y_i &= 10X_{i1} + 5X_{i2}^2 + X_{i3}^3 + \e \label{linear2},
\end{align}
where $\e \sim N(0,\sigma^2)$ and $\sigma \in \{1,5,10\}$. Root mean  square error on the truth is reported, averaged over 50 runs, for lasso, local linear regression, BART, random forests, adaptive random forests, and local linear forests. In the simple linear case in equation \ref{linear1}, we see that lasso outperforms the other methods, as we would expect; in the polynomial given in equation \ref{linear2}, local linear regression performs the best, followed by BART ($\sigma=1$ case) and local linear forests ($\sigma= 5, 10$ cases).

\begin{table}[h]
\begin{center}
\footnotesize
\begin{tabular}{@{}lllllllll@{}}
\toprule \toprule
Setup & $\sigma$ && lasso & LLR & BART & RF & LLF  \\
\midrule
Equation \ref{linear1} & 1 && {\bf 0.12} & 0.15 & 0.48 & 0.73  & 0.22 \\
 & 5 && {\bf 0.39} & 0.92 & 1.27 & 1.25 &  0.96\\
  & 10 && {\bf 0.70} & 1.70 & 2.37 & 1.76 & 1.56  \\
\midrule\midrule
Equation \ref{linear2} & 1 && 1.55 & {\bf 0.22} & 0.50 & 0.86 & 0.69 \\
 & 5 && 1.55 & {\bf 0.92} & 1.31 & 1.32 & 1.28 \\
  & 10 && 1.66 & {\bf 1.44} & 1.83 & 1.70 & 1.68 \\
\bottomrule
\bottomrule 
\end{tabular}
\end{center}
\caption{Root Mean Square Error from simulations on equations \ref{linear1} and \ref{linear2} on lasso, local linear regression (LLR), BART, random forests, adaptive random forests, and local linear forests. We vary error variance $\sigma$ from 1 to 10 and fix $n=600, d=3$. All errors are averaged over 50 runs, and minimizing errors are in bold.}\label{linear-table}
\end{table}

\subsection{Proof of Theorem 1}
\label{sec:pf}

Throughout this proof, we use the notation $M_\lambda$ established in (19), and
shorthand $Y_i = \mu(X_i) + \e_i$.
Define the diameter (and corresponding radius) of a tree leaf as the length of the longest line segment that can fit completely inside of the leaf. 
Thanks to our assumed uniform bound on the second derivative of $\mu(\cdot)$,
a Taylor expansion of around $\mu({x})$ around $x_0$ yields the following decomposition starting from (5):
\begin{equation}\label{decomposition}
\begin{split}
&\hat{\mu}({x_0}) = e_1^T M_{\lambda}^{-1}  \sum_{i=1}^n\begin{pmatrix} 1 \\ X_i- {x_0} \end{pmatrix}  \alpha_i({x_0})  Y_i = \mu({x_0}) + \hat{\gamma}_n({x_0}) +  Q({x_0}) +  O\left(\bar{R^2}\right), \\
&\hat{\gamma}_n(x_0) = e_1^T M_{\lambda}^{-1} \sum_{i=1}^n    \begin{pmatrix} 1 \\ X_i- {x_0} \end{pmatrix} \alpha_i({x_0}) \e_i, \\
&Q(x_0) =e_1^TM_{\lambda}^{-1} \sum_{i=1}^n  \begin{pmatrix} 1 \\ X_i- {x_0} \end{pmatrix}   \alpha_i({x_0}) \p{\nabla \mu({x_0}) \cdot \p{X_i- {x_0}}}, 
\end{split}
\end{equation}
where $\bar{R^2}$ is the average squared radius of leaves $T_b$ in the forest.
In other words, we have decomposed our forest into a variance term $\hat{\gamma}_n({x_0})$,
a regularization bias term $Q({x_0})$, and a curvature bias term that's bounded on the order of $\bar{R^2}$.
Our main goal is to show that we can approximate $\hat{\gamma}_n({x_0})$ via an (infeasible) regression forest,
while the remaining terms are lower order.
For simplicity, moving forward we will write $\alpha_i({x_0}) = \alpha_i$, dropping the written dependence on ${x_0}$.

\paragraph{Curvature bias}

To control the curvature bias, we need to control the radius $R_{T_b}$ of a typical leaf containing ${x_0}$.
To do so, we use the following bound.
Recall that $X_1, \dots, X_s \sim U([0,1]^d)$ independently, and that $T_b$ is a regular, random-split tree.
By Lemma 2 of \citet*{wager2017}, we then see that for any $0<\eta<1$ and for large enough $s$,
\begin{equation}
\label{eq:WAlem2}
\mathbb{P}\left[\text{diam}_j(L({x_0})) \ge \left(\frac{s}{2k-1}\right)^{-\frac{0.99 (1-\eta)\log((1-\omega)^{-1})}{\log(\omega^{-1})} \frac{\pi}{d}} \right] \le \left(\frac{s}{2k-1}\right)^{-\frac{\eta^2}{2} \frac{1}{\log(\omega^{-1})} \frac{\pi}{d}},
\end{equation}
where $k$ is (fixed) the tree-depth parameter from Assumption 1.
We start by applying \eqref{eq:WAlem2} with $\eta = 0.49$, and note that $0.99  (1 - 49) > 0.5$ and
$0.49^2 / 2 / \log(1/0.8) > 0.53$, meaning that for all $\omega \leq 0.2$,
\begin{equation}
\label{diameter-bound}
\prob\left(\text{diam}_j(L({x_0})) \ge r_s \right) \le r_s^{1.06}, \ \ \ \ \ r_s =  s^{-\frac{1}{2}\frac{\log((1-\omega)^{-1})}{\log(\omega^{-1})} \frac{\pi}{d}}.
\end{equation}
This suggests that most leaves should have radius bounded on the order of $r_s$.
To get a useful bound on the second moment of leaf radii via $\bar{R^2}$, though, we need to use chaining: Setting $\eta = 0.71$,
we find that
\begin{equation*}
\prob\left(\text{diam}_j(L({x_0})) \ge r_s^{0.57} \right) \le r_s^{2.2}.
\end{equation*}
Then, applying Markov's inequality twice, we see that $\bar{R^2} = O_p(r_s^2)$.

\paragraph{Regularization bias}

The term $Q(x_0)$ in \eqref{decomposition} has more intricate behavior. We note that, if we had no regularization
at all, then the local linear correction would perfectly adjust for the slope of $\mu(\cdot)$ and $x_0$, and so we would have
$Q(x_0) = 0$; unfortunately, however, we need positive regularization in other parts of the proof so we cannot
directly use this fact. Conversely, as $\lambda \rightarrow \infty$, the local linear forest becomes a regression forest, and
$Q(x_0)$ becomes a bias term on the order of $\bar{R}$; and this was the dominant bias term in the analysis of \citet*{wager2017}.

The derivation shows that, given a reasonable amount of regularization $0 < \lambda < \infty$,
the term $Q(x_0)$ is non-zero but still much smaller than $\bar{R}$. Recall our notation $\Delta_i$
denoting a $p+1$-dimensional vector consisting of a 1 stacked with $X_i - x_0$, and let $v = (0, \, \nabla \mu(x_0))$.
Then, writing $\Delta$ for the matrix with rows $\Delta_i$ and plugging in the expression 19
for $M_\lambda$, we see that
\begin{align*}
Q(x_0) &= e_1' \p{\Delta' A \Delta+ \lambda J}^{-1} \Delta' A \Delta v \\
&= -e_1' \p{\Delta' A \Delta+ \lambda J}^{-1} \lambda J v \\
&= -\lambda e_1' \p{\Delta' A \Delta+ \lambda J}^{-1} v \\
&= \lambda \p{1 - d_\alpha' \p{S_\alpha + \lambda I}^{-1} d_\alpha}^{-1} d_\alpha' \p{S_\alpha + \lambda I}^{-1} \nabla \mu(x_0),
\end{align*}
where the last line followed from the Schur formula, with notation $d_\alpha = \sum_{i = 1}^n \alpha_i (X_i - x_0)$ and
$S_\alpha = \sum_{i = 1}^n \alpha_i (X_i - x_0)^{\otimes 2}$ as used in Assumption 3. 
We now make some observations. First, by Assumption 3
$$ \p{1 - d_\alpha' \p{S_\alpha + \lambda I}^{-1} d_\alpha}^{-1} = O_p(1) $$
is of constant order in probability.
Second, by Cauchy-Schwarz, 
\begin{align*}
d_\alpha' \p{S_\alpha + \lambda I}^{-1} \nabla \mu(x_0)
&\leq \sqrt{d_\alpha \p{S_\alpha + \lambda I}^{-1} d_\alpha} \sqrt{\nabla \mu(x_0)' \p{S_\alpha + \lambda I}^{-1} \nabla \mu(x_0)} \\
&\leq \lambda^{-1/2}  \Norm{\nabla \mu(x_0)}_2,
\end{align*}
noting that $d_\alpha' S_\alpha^{-1} d_\alpha \leq 1$ by Jensen's inequality. Combining all these facts
together, we find that $Q(x_0) = O_p(\sqrt{\lambda})$.

\paragraph{The variance term}

Finally, we turn to the variance term \smash{$\hat{\gamma}_n({x_0})$}.
To do so, our main task is to couple \smash{$\hat{\gamma}_n$} with an approximation \smash{$\tilde{\gamma}_n$}, defined as 
\begin{equation}
\label{eq:oracle}
\tilde{\gamma}_n({x_0}) = \sum_{i=1}^n \alpha_i \tilde{Y}_i, \text{~~where~~} \tilde{Y}_i =  e_1^T \E[M_{\lambda}]^{-1} \begin{pmatrix} 1 \\ X_i- {x_0} \end{pmatrix} \e_i.
\end{equation}
Now, we note that \smash{$\tilde{Y}_i$} is independent of $\alpha_i$ conditionally on $X_i$
(because the problematic associations discussed at the beginning of Section 4 were mediated by $M_\lambda$),
and so \smash{$\tilde{\gamma}_n({x_0})$} is just the prediction made by a ``regression forest'' with outcome $\tilde{Y}_i$.
Consequently \smash{$\tilde{\gamma}_n$} can be characterized via standard tools used to study random forests.

We sketch out an argument below, based on the fact that $M_{\lambda}$ concentrates around its expectation.
Following the line of argumentation in \citet{wager2017}, we see that $M_{\lambda}$ is a $U$-statistic with
kernel size $s$. Moreover, by \eqref{diameter-bound}, we see that the stochastic fluctuations of the terms
forming $M_\lambda$ are of order $r_s^2$. Thus, we can use concentration inequalities for $U$-statistics
following \citet*{hoeffding1963} to verify that (to use this concentration inequality, we need to perform several steps of
chaining following \eqref{diameter-bound}, going up to $\eta = 0.98$)
\begin{equation}
\label{eq:Mlambda_conc}
\Norm{M_\lambda - \E[M_\lambda]}_\infty = O_p\p{r_s^2 \sqrt{s/n}}.
\end{equation}
Next, note that
\begin{equation}
\label{eq:gerr}
\hat{\gamma}_n({x_0}) - \tilde{\gamma}_n({x_0})
= e_1 \p{M_\lambda^{-1} - \E[M_\lambda]^{-1}} \Delta' A \e.
\end{equation}
Thus, because $\e$ is independent of all other terms in \eqref{eq:gerr}, we see that
the discrepancy between $\hat{\gamma}_n({x_0})$ and $\tilde{\gamma}_n({x_0})$ is bounded
on the order of $\Norm{e_1 \p{M_\lambda^{-1} - \E[M_\lambda]^{-1}} \Delta' A}_2$; an
application of the Schur formula together with \eqref{eq:Mlambda_conc} then implies that
\begin{equation}
\label{eq:gerr2}
\hat{\gamma}_n({x_0}) - \tilde{\gamma}_n({x_0})
= O_p\p{\lambda^{-2} r_s^4 \, s /n}
\end{equation}
for all $\lambda \gg r_s^2 \sqrt{s/n}$.

\paragraph{Wrapping up}

We are now ready to put everything together. Given everything we've seen so far, we've established that
\begin{equation*}
\hat{\mu}({x_0}) - \mu(x_0) =  \tilde{\gamma}({x_0}) + O_p\p{r_s^2 + \sqrt{\lambda} + \lambda^{-2} r_s^4 \frac{s}{n}}
\end{equation*}
for all $\lambda \gg r_s^2 \sqrt{s/n}$. Thus, setting $\lambda = \Theta(r_s^{1.98} \sqrt[4]{s/n})$ as in (17), we get
\begin{equation*}
\hat{\mu}({x_0}) - \mu(x_0) =  \tilde{\gamma}({x_0}) + O_p\p{r_s^2 + r_s^{0.99} \sqrt[8]{s/n} + r_s^{0.04}  \sqrt{s/n}}.
\end{equation*}
Now, recall that we have chose $s = n^{\beta}$ for some $\beta \geq \beta_{\min}$, meaning that
\begin{align*}
\sqrt[3/8]{s/n} = s^{\frac{3(1 - \beta^{-1})}{8}} 
\geq s^{-\frac{3 \times 1.3}{8} \frac{\log((1-\omega)^{-1})}{\log(\omega^{-1})} \frac{\pi}{d}} 
\gg r_s^{0.99},
\end{align*}
and so the above expression simplifies to
\begin{equation}
\label{eq:errrr}
\hat{\mu}({x_0}) - \mu(x_0) =  \tilde{\gamma}({x_0}) + o_p\p{\sqrt{s/n}}.
\end{equation}
It remains to show that $\tilde{\gamma}({x_0})$ is asymptotically centered and Gaussian with errors on the
scale of $\sqrt{s/n}$, meaning that $\tilde{\gamma}({x_0})$ is in fact the dominant error term in $\hat{\mu}({x_0})$.

But now, recall that $\tilde{\gamma}({x_0})$ is simply a regression forest with outcome $\tilde{Y}_i$. Thus,
Theorem 8 of \citet{wager2017} directly implies that there is sequence $\sigma_n({x_0}) \to 0$ such that 
\begin{equation}
\label{eq:pfclt}
\frac{\tilde{\gamma}_n({x_0})}{\sigma_n({x_0})} \Rightarrow \mathcal{N}(0,1);
\end{equation}
here, we used the fact that the $\e_i$ are all mean-zero conditionally on the tree construction, and so $\E[\tilde{\gamma}({x_0})] = 0$.
Finally, from Theorem 5 of \citet{wager2017}, we see that
$\sigma_n({x_0}) = \sqrt{s/n} \operatorname{polylog}(s)$, and we note that our above argument in fact established
a polynomial gap between the error term in \eqref{eq:errrr} and $\sqrt{s/n}$. Thus \eqref{eq:pfclt} in fact captures
the dominant error term of our estimator.

\end{document}